\documentclass[11pt]{article}

\usepackage[english]{babel}

\usepackage[letterpaper,top=2cm,bottom=2cm,left=3cm,right=3cm,marginparwidth=1.75cm]{geometry}

\usepackage{url}
\usepackage{amsmath}
\usepackage{graphicx}
\usepackage[colorlinks=true, allcolors=blue]{hyperref}

\usepackage{subcaption}
\usepackage{enumitem}
\usepackage{algorithm}
\usepackage{algpseudocode}
\usepackage{algorithmicx}

\newcommand{\ie}{\emph{i.e.}}

\newcommand\blfootnote[1]{%
  \begingroup
  \renewcommand\thefootnote{}\footnote{#1}%
  \addtocounter{footnote}{-1}%
  \endgroup
}

\usepackage{authblk}  

\title{MR-EEGWaveNet: Multiresolutional EEGWaveNet for Seizure Detection from Long EEG Recordings}

\author[1]{Kazi Mahmudul Hassan}
\author[2,3,4]{Xuyang Zhao}
\author[5]{Hidenori Sugano}
\author[1]{Toshihisa Tanaka\thanks{Corresponding author: tanakat@cc.tuat.ac.jp}}

\affil[1]{Department of Electronic and Information Engineering, Tokyo University of Agriculture and Technology, Tokyo, Japan}
\affil[2]{Medical Science Data-driven Mathematics Team, RIKEN Center for Interdisciplinary Theoretical and Mathematical Sciences, Kanagawa, Japan}
\affil[3]{Medical Data Mathematical Reasoning Special Team, RIKEN Center for Integrative Medical Sciences, Kanagawa, Japan}
\affil[4]{Department of Artificial Intelligence Medicine, Chiba University, Chiba, Japan}
\affil[5]{Department of Neurosurgery, Juntendo University School of Medicine, Tokyo, Japan}

\newbox\keywbox
\setbox\keywbox=\hbox{\bfseries Keywords:}%
\newcommand\keywords{%
\noindent\rule{\wd\keywbox}{0.25pt}\\\textbf{Keywords:}\ }
\newcommand{\creditline}[1]{\par\vspace{0.5em}\noindent\textbf{Acknowledgment:} #1\par}

\begin{document}
\maketitle
\begin{abstract}
Feature engineering for generalized seizure detection models remains a significant challenge. Recently proposed models show variable performance depending on the training data and remain ineffective at accurately distinguishing artifacts from seizure data. In this study, we propose a novel end-to-end model, ``Multiresolutional EEGWaveNet (MR-EEGWaveNet),'' which efficiently distinguishes seizure events from background electroencephalogram (EEG) and artifacts/noise by capturing both temporal dependencies across different time frames and spatial relationships between channels. The model has three modules: convolution, feature extraction, and predictor. The convolution module extracts features through depth-wise and spatio-temporal convolution. The feature extraction module individually reduces the feature dimension extracted from EEG segments and their sub-segments. Subsequently, the extracted features are concatenated into a single vector for classification using a fully connected classifier called the predictor module. In addition, an anomaly score-based post-classification processing technique is introduced to reduce the false-positive rates of the model. Experimental results are reported and analyzed using different parameter settings and datasets (Siena (public) and Juntendo (private)). The proposed MR-EEGWaveNet significantly outperformed the conventional non-multiresolution approach, improving the F1 scores from $0.177$ to $0.336$ on Siena and $0.327$ to $0.488$ on Juntendo, with precision gains of $15.9\%$ and $20.62\%$, respectively.
\end{abstract}

\creditline{This work was supported in part by JSPS KAKENHI 23H00548. We would like to thank Editage (www.editage.jp) for English language editing.}

\section{Introduction}
\label{sec:intro}
Epilepsy is a group of neurological disorders, characterized by unprovoked and recurrent seizures in a patient over a certain period \cite{hassan_epilepsy_CW_2019}. A seizure is an outcome of abnormal activity in the brain. This abnormal activity can affect the entire brain or be confined to a particular brain region, determining the types of seizures \cite{scheffer_span_2017}. Seizures can be caused by genetic disorders or brain injury; however, their underlying reasons often remain unknown \cite{fordington_review_2020}. The physiological and psychological effects of epilepsy in a patient can be reduced with an early and affordable diagnosis. 

The electroencephalogram (EEG) is an essential tool for the diagnosis of epilepsy, as it captures electrical signals from the brain through electrodes \cite{Baumgartner_scalpEEG_2018}. In medical practice, scalp EEG is widely used for the primary inspection of seizure events. This process is non-invasive, affordable, and easily set up by practitioners. In addition, as seizures are rare events, their detection in EEG requires extensive data collection, often spanning hours, or days. Detecting seizures through an EEG is an important but cumbersome process for clinicians \cite{moutonnet_clinical_2024_review}. It requires higher-level expertise to distinguish seizure events from background EEG. Therefore, automatic seizure detection has been in high demand among medical professionals. 

With advancement in technology and the availability of EEG data, numerous attempts have been made over the past few decades to automate the seizure detection system \cite{moutonnet_clinical_2024_review}. EEG data are inherently statistically variable across patients, making the feature engineering task quite challenging. Additionally, the nonlinear, non-stationary, and subject-dependent nature of EEG, including a heterogeneous recording environment, makes the task complex, especially in the case of scalp EEG recordings. Recent advancements in deep learning methods have inspired many researchers to develop models for detecting seizure events more efficiently \cite{Zhang_seizure_review_2024, qiu_lightseizurenet_2023, poorani_deep_2023, raab_xai4eeg_2023, ibrahim_deeplearningbased_2022, zhang_scheme_2024, zhao_residual_2024}. Most existing models are based on convolutional neural networks (CNNs). However, several limitations exist in these studies, particularly regarding the use of hybrid architectures, validation techniques, and validation datasets. In hybrid architectures, feature extraction often partially depends on hand-crafted feature engineering. This approach may limit the generalizability of the model and fully utilize the deep learning representation capabilities. Furthermore, the K-fold cross-validation technique is a common validation approach to evaluate model performance. In epileptic datasets, considering the limited number of subjects and practical scenario, the leave-one-subject-out (LOSO) validation approach is more appropriate than K-fold cross-validation \cite{kunjan_LOS0_2021, Wong_LOSO_2015}. In K-fold cross-validation, there is a high possibility that data from the same subject may appear in both the training and the testing sets, potentially leading to an overestimation of the model's performance. Additionally, some publicly available EEG datasets \cite{bonn_dataset_2001}, which are preprocessed to be noise-free and artifact-free, contain noninvasive and invasive EEGs together. When a model is validated on such datasets, it can significantly overestimate its true effectiveness in real-world scenarios. An end-to-end model called ``EEGWaveNet'' has demonstrated promising performance in the seizure detection task \cite{thuwajit_eegwavenet:_2022}. However, the performance of the model significantly depends on the training strategy and requires preprocessing to reduce the amount of noise or artifacts in the dataset. In contrast, the recording techniques for scalp EEG recordings are highly noisy or artifact-prone \cite{Kaya_summary_EEG_artifact_2021}. This can lead to frequent false seizure detections, undermining the model's reliability.

In this study, we propose an extended version of ``EEGWaveNet,'' named ``MR-EEGWaveNet'', which improves the performance of the original ``EEGWaveNet'' model by reducing the false detection of seizures for long EEG recordings. A key challenge for deep models in seizure detection lies in achieving balanced performance on highly imbalanced datasets, where accurate identification of the minority seizure class is critical. In contrast to the EEGWaveNet model, which uses fixed segment length, the MR-EEGWaveNet extracts features from the EEG segment and its sub-segments (multiple segments within the segment) and calculates their features for the classification task. The experimental results indicate that the MR-EEGWaveNet can obtain an optimal trade-off point of performance between recall and specificity, improving the precision score. We experimented with different parameter settings and performed a comparative analysis.   

The major contributions of this paper are as follows: 
\begin{enumerate}[label=\roman*.]
    \item Proposed an extended version of the ``EEGWaveNet'' considering multiresolutional analysis in the feature extraction module. 
    \item  A novel post-classification processing technique based on an anomaly score calculated by a state-of-the-art method further minimizes the misdiagnosis of the nonseizure EEG segment as seizure.  
\end{enumerate}
The performance of the MR-EEGWaveNet is evaluated with a public dataset (``Siena'') and a private dataset (``Juntendo''). While direct comparison with existing methods is difficult due to differing parameters, evaluation strategies, and datasets, we performed a comparative analysis with recent state-of-the-art models. The results indicate that the MR-EEGWaveNet performs better in several key evaluation metrics.

The paper is organized as follows.
Section \ref{sec:previous_works} presents recent and relevant studies in the seizure detection task.
Section \ref{sec:dataset} briefly discusses the experimental datasets.
Section \ref{sec:proposed_model} contains the details of the proposed model.
Sections \ref{sec:experiment}, \ref{sec:results}, and \ref{sec:discussion} describe the experimental setup, report the results, and discuss the findings, respectively.
Finally, Section \ref{sec:conclusion} provides concluding remarks.

\section{Previous Works}
\label{sec:previous_works}
Several studies have previously explored potential solutions to the seizure detection problem \cite{acharya_deep_2018, ullah_automated_2018, liu_patient-independent_2022, thuwajit_eegwavenet:_2022}. They proposed several models and extracted features to distinguish seizure and nonseizure events from long recordings of EEG. In EEG research, the calculated features used to separate brain activity are commonly categorized into three groups: complexity, continuity, and connectivity \cite{saba_sadiya_unsuper_artifact_2021}. In epilepsy research, complexity features (such as entropy and fractal-based methods) measure the irregularity in the data to detect seizure activity in EEG. Recently, different types of entropy have been proposed for the seizure detection task \cite{molla_hassan_graph_sz_det_sensor2020, Akter_multiband_2020}. Continuity-based features, such as band power ($\delta$, $\theta$, $\alpha$, $\beta$, and $\gamma$), median frequency, diffuse slowing, and burst suppression \cite{stern2005atlas_EEG_pattern}, calculate the stability and persistence of EEG signals over time. Connectivity features, such as coherence, phase-locking value, and mutual information, assess the relationship and interaction between EEG channels that cover different regions of the brain \cite{saba_sadiya_unsuper_artifact_2021}. Furthermore, a combination of time-, frequency-, and entropy-based features has been applied with a random forest classifier to detect seizures in EEG \cite{Sigsgaard2023_SZ_detection}.  

Although these features have been used in several studies, challenges and limitations, such as sensitivity to noise, longer time window requirements, limited temporal resolution, capture of linear relationships in EEG, and complex interpretation of brain connectivity, are encountered in the practical domain \cite{saba_sadiya_unsuper_artifact_2021}. In addition, artifacts are common in scalp EEG recordings that reduce the performance of seizure detection models. The statistical properties of artifact and seizure events are similar, and their probability of simultaneous occurrence is relatively high \cite{Kaya_summary_EEG_artifact_2021}. To address this issue, the Riemannian manifold-based methods have recently gained popularity in the brain--computer interfacing domain. The Riemannian manifold is a nonlinear space that can be constructed from EEG \cite{Yger2017_riemann_review, yamamoto_Riemann_2020}. The Riemannian features are calculated through covariance matrices generated from EEG signals \cite{orihara_hassan_active_sel_2023}. These features can be used effectively to detect anomalies in EEG and separate artifacts and seizures \cite{hassan_ano_det_ICASSP2024, hassan_rimann_ATSIP_2024}.

However, the nonlinear and nonstationary nature of EEG presents a significant challenge in extracting features that consistently perform well over time. In general, EEG signals exhibit substantial inter-subject variability, and the features extracted from them are often highly subject-specific. In recent years, researchers have been actively developing data-driven approaches, such as end-to-end deep learning models to overcome feature engineering challenges in EEG \cite{liu_patient-independent_2022,thuwajit_eegwavenet:_2022}. Convolutional neural networks (CNN) and bidirectional long short-term memory are well-known models that are used together to detect seizures, as proposed in \cite{liu_patient-independent_2022}. Using a discrete wavelet transform, the EEG signals were decomposed into subbands, and the features were extracted using CNNs. Subsequently, the false detection rate and sensitivity were improved by smoothing and collar techniques. By transforming EEG into plot images, a CNN-based model for seizure detection was designed in \cite{CNN_sz_det_EMAMI2019}. The model was improved with the addition of a patient-specific autoencoder, which introduced a new label, ``nonseizure-but-abnormal,'' to the existing binary labels (``seizure'' and ``nonseizure'') \cite{takahashi_AE_CNN_2020}. Recently, graph-based analysis has also gained popularity in seizure detection. In \cite{Shen_googleNet_sz_2024}, the short-time Fourier transform (STFT) of EEG signals was transformed into graph data and then classified with Google-Net CNN models. In \cite{JIBON_linearGCNN_2023}, the EEG signal was first transformed into a spectrogram using the Stockwell transform. A Linear Graph Convolution Network (GCN) was used for feature selection, followed by DenseNet-based classification. ``EEGWaveNet'' is a deep learning model to process EEG signals introduced in \cite{thuwajit_eegwavenet:_2022}. It can capture both time-related and channel-related features that are suitable for classifying EEG signals compared to other models. However, implementing deep learning-based models in the medical domain (specifically seizure detection) has faced several challenges. Due to the complex and subject-dependent nature of seizures, these techniques have faced challenges, such as extreme data imbalance, complex interpretation, and the need for large amounts of training data to fine-tune hyperparameters.

EEGNet, proposed in \cite{EEGNet_lawhern2018}, is a widely adopted model specifically developed for brain-computer interface tasks. Through the use of temporal, depthwise, and separable convolutions, the model captures key frequency elements, spatial features specific to those frequencies, and temporal patterns.
In contrast, ``EEG Conformer'' adopts a compact convolution-attention architecture \cite{Song_EEG_conformer_2023}. It captures short-term temporal and spatial features using convolutional layers, while self-attention models long-term global temporal dependencies. This design enables a unified and robust framework for EEG classification.

The success of large language models in other domains has also inspired researchers to develop EEG-based foundation models recently \cite{Yang_BIOT_2023, Jiang_LaBraM_2024, wang2025cbramod}. 
Biosignal Transformer (BIOT) \cite{Yang_BIOT_2023} preprocesses biosignals into tokenized sequences and employs a linear transformer to capture temporal interactions. Its encoder supports supervised learning, pre-training, and fine-tuning on both complete and incomplete data across various biosignal tasks.
In \cite{Jiang_LaBraM_2024}, the authors proposed a Large Brain Model (LaBraM) to overcome challenges such as varying electrode counts, low signal-to-noise ratios, and more. The LaBraM model segments EEG signals into fixed-length patches, applies temporal encoding, adds temporal and spatial embeddings, and then processes the sequence with a Transformer encoder using patch-wise attention for final feature extraction.
Recently, a brain foundation model named  Criss-Cross Brain Foundation Model (CBraMod) has been proposed \cite{wang2025cbramod}. It employs a criss-cross transformer backbone to independently model spatial and temporal dependencies in EEG signals via parallel attention. Additionally, It uses an asymmetric conditional positional encoding scheme that encodes EEG patch positions and easily adapts to diverse EEG formats.

In the case of epileptic seizure detection tasks, post-classification processing is a common approach to improve the performance of the system \cite{moutonnet_clinical_2024_review, post_process_khalkhali_low_2021, post_process_wei_2020}. In \cite{post_process_khalkhali_low_2021}, the authors adopted a post-processing approach based on multiple thresholds. They classified an event as a seizure if its probability exceeded the seizure threshold or if a background event was shorter than the minimum acceptable duration. In addition, seizures with a duration shorter than the minimum threshold were treated as background activity. In \cite{post_process_wei_2020}, seizures detected using an XGboost-based method with intervals shorter than $2$ s were grouped and reclassified as nonseizure events if their duration was less than $15$ s.

Detecting seizures in long-term scalp EEG recordings is particularly challenging due to the frequent appearance of abnormal activities and artifacts. Therefore, special care and checking are required both during and after EEG recording. The selection of window length plays an important role in seizure detection; more specifically, in deep learning-based methods, where techniques, such as convolution are used \cite{CNN_sz_det_EMAMI2019}. Larger windows are suitable for extracting statistical properties, whereas a shorter windows provide more temporal information. Seizures are a complex and subject-dependent process. Certain seizures show consistent characteristics within and between consecutive time frames, while others have evolution, that is, their seizure characteristics change over time \cite{meritamlarsen_duration_2023}. In addition, certain types of seizures have results, such as blinking in the eye, rapid movement in the eye, muscle contractions, stiffening, rhythmic jerking, lip smacking, hand movements, or chewing \cite{TUSZ_2016}. Those artifacts originate outside the brain and severely affect the pattern or shape of the EEG signal. 
Thus, the context of those artifacts is important to consider when designing a seizure detection model. However, different types of seizures and artifacts require varying window lengths to be effectively represented and distinguished by the detection model. Certain artifacts and seizures are better characterized with short windows for fine temporal details, while others require longer windows to extract broader patterns. For example, a high-amplitude spike artifact within a $10$-second window may generate features that are indistinguishable from those of a seizure event if adjacent temporal dynamics are not considered.

A potential solution to the aforementioned problem is the multiresolutional feature analysis using deep learning-based models. Among recently proposed end-to-end models, the simplicity and potential of the EEGWaveNet model in seizure detection inspire us to further investigate it for possible performance improvement through architectural enhancements. In this study, we propose a new architecture based on the EEGWaveNet model, named ``MR-EEGWaveNet,'' designed to improve the performance of the existing model. The model consists of three main modules: convolution, feature extraction, and predictor. In contrast to EEGWaveNet, additional features are calculated at multiple resolutions from the sub-segments of the input EEG segment. This model explores the relationship among different parts of the EEG signal through multi-scale and spatio-temporal convolutions, including pooling operations. For each segment and its sub-segment, the feature extraction module extracted features from each ``Spatio-temporal Convolution'' module and concatenated them. Finally, the features from the segments and sub-segments are concatenated together to feed into the predictor module for the classification. This additional feature extraction from multiple resolutions enhances the ability to capture fine details of the short-time event while preserving their broader context. In addition, we introduced an anomaly score-based post-classification processing technique to improve the performance of the classification model.

\begin{figure}[t!]
    \centering
     \begin{subfigure}{0.8\textwidth}
        \centering
        \centerline{\includegraphics[width=1.0\linewidth]{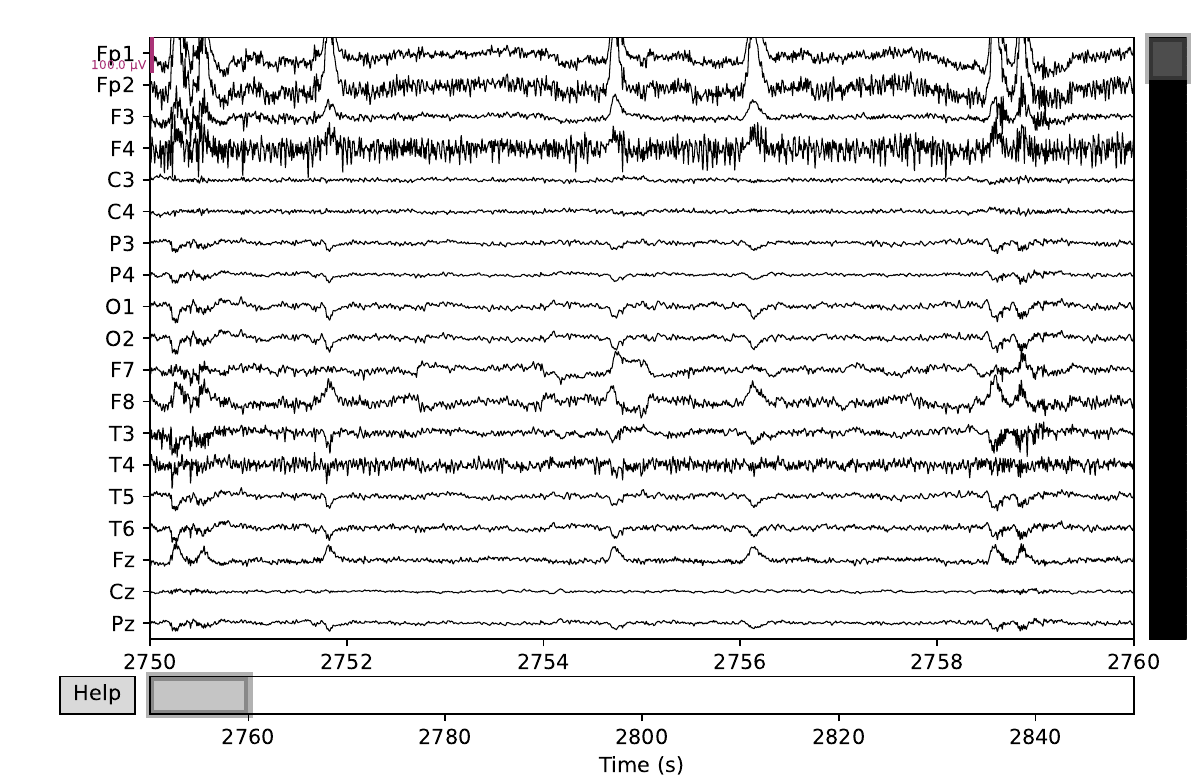}}
        \caption{Nonseizure EEG}
        \label{fig:ns_eeg}
	\end{subfigure}
 
   \vspace{1em}
 
    \begin{subfigure}{0.8\textwidth}
        \centering
        \centerline{\includegraphics[width=1.0\linewidth]{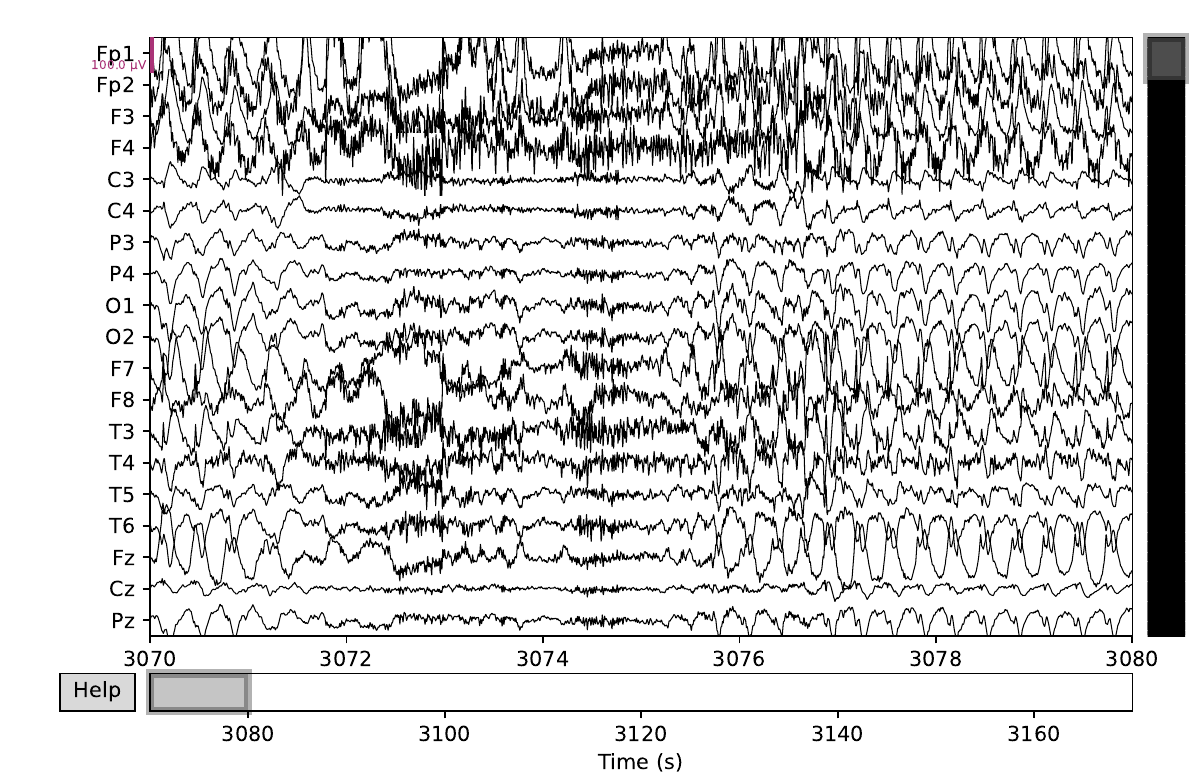}}
        \caption{Seizure EEG }
        \label{fig:sz_eeg}
	\end{subfigure}
 
\caption{$10$-seconds long (a) Nonseizure, (b) Seizure EEG segments from Juntendo dataset.}
\label{fig:EEG_sz_ns_plot}
\end{figure}

\section{Dataset}
\label{sec:dataset}

\subsection{Public Dataset}
\label{sec:siena_dataset}
A publicly available EEG dataset was employed. The dataset was recorded at the Unit of Neurology and Neurophysiology of the University of Siena \cite{detti_eeg_2020}; hereafter, it will be referred to as ``Siena''. The dataset comprised recordings of 14 patients, including eight males (ages 25--71) and six females (ages 20--58). The International 10-20 standard was used for EEG recordings with the $512$ Hz sampling frequency. The EEG recordings were downsampled to 500 Hz for the experiment. The dataset was diagnosed and labeled (seizure events) by an expert clinician according to the International League Against Epilepsy criteria \cite{detti_siena_nodate}. This study was conducted using EEG recordings from 19 commonly used channels: Fp1, Fp2, F7, F8, F3, F4, T3, T4, T5, T6, C3, C4, P3, P4, Fz, Cz, Pz, O1, and O2.  Detailed information and statistics of the dataset are summarized in Table \ref{tab:siena-data-table}. The dataset is publicly available and can be accessed in \cite{detti_siena_nodate}.

\begin{table*}[t!]
\centering
\caption{Description of the ``Siena'' EEG dataset. (Pt.: Patient ID; IAS: Focal onset impaired awareness, WIAS: Focal onset without impaired awareness, FBTC: Focal to bilateral tonic-clonic; Loc.: Localization, T: Temporal lobe, F: Frontal lobe; Lat.: Lateralization)}
\label{tab:siena-data-table}

\scalebox{0.80}
{
\begin{tabular}{l|c|c|c|c|c|c|c|c|c}
\hline
    \textbf{Pt.} 
    & \textbf{Gender} 
    & \begin{tabular}[c]{@{}l@{}}\textbf{Age} \\ \textbf{(year)}\end{tabular} 
    & \begin{tabular}[c]{@{}l@{}}\textbf{Seizure} \\ \textbf{types}\end{tabular} 
    & \textbf{Loc.} 
    & \textbf{Lat.}
    & \begin{tabular}[c]{@{}l@{}}\textbf{Record}\\\textbf{Length(hr.)}\end{tabular} 
    & \begin{tabular}[c]{@{}l@{}}\textbf{Num. of} \\ \textbf{Seizures}\end{tabular} 
    & \begin{tabular}[c]{@{}l@{}}\textbf{Seizure} \\ \textbf{time (sec.)}\end{tabular} 
    & \begin{tabular}[c]{@{}l@{}}\textbf{Seizure} \\ \textbf{ratio} (\%)\end{tabular} \\
    
\hline
    PN00   & Male    & 55  & IAS     & T     & Right        & 3.20     & 5   & 325     & 2.821  \\\hline
    PN01   & Male    & 46  & IAS     & T     & Left         & 13.48    & 2   & 128     & 0.264  \\\hline
    PN03   & Male    & 54  & IAS     & T     & Right        & 24.22    & 2   & 244     & 0.280  \\\hline
    PN05   & Female  & 51  & IAS     & T     & Left         & 6.02     & 3   & 104     & 0.480  \\\hline
    PN06   & Male    & 36  & IAS     & T     & Left         & 12.05    & 5   & 282     & 0.650  \\\hline
    PN07   & Female  & 20  & IAS     & T     & Left         & 8.73     & 1   & 62      & 0.197  \\\hline
    PN09   & Female  & 27  & IAS     & T     & Left         & 6.83     & 3   & 203     & 0.825  \\\hline
    PN10   & Male    & 25  & FBTC    & F     & Bilateral    & 18.68    & 10  & 365     & 0.543  \\\hline
    PN11   & Female  & 58  & IAS     & T     & Right        & 2.40     & 1   & 55      & 0.637  \\\hline
    PN12   & Male    & 71  & IAS     & T     & Left         & 6.07     & 4   & 290     & 1.328  \\\hline
    PN13   & Female  & 34  & IAS     & T     & Left         & 8.63     & 3   & 264     & 0.849  \\\hline
    PN14   & Male    & 49  & WIAS    & T     & Left         & 20.43    & 4   & 163     & 0.222  \\\hline
    PN16   & Female  & 41  & IAS     & T     & Left         & 4.85     & 2   & 230     & 1.317  \\\hline
    PN17   & Male    & 42  & IAS     & T     & Right        & 5.10     & 2   & 153     & 0.833  \\\hline
\end{tabular}
}
\end{table*}

\subsection{Private Dataset}
\label{sec:junt_dataset}
In this study, a private EEG dataset recorded at Juntendo University Hospital in Japan was employed. The ethics committees of Juntendo University Hospital and Tokyo University of Agriculture and Technology approved the research. The dataset consists of scalp EEG recordings from multiple epileptic patients, with several hours of data collected for each. Specifically, 21 patients' data (thirteen males and eight females), ranging in age from 1 to 42 years, were used in the experiment. The electrode placement followed the International 10-20 system, with a sampling frequency of 500 Hz. The number of selected channels was 19, the same as the Siena EEG dataset. The dataset contains different types of seizures, such as generalized, motor, nonmotor, focal, non-focal, tonic-clonic, behavioral arrest, among others. The brief data statistics are shown in Table \ref{tab:juntendo-data-table}. The $10$-s long nonseizure and seizure EEG segment from the dataset is shown in Figure \ref{fig:EEG_sz_ns_plot}.

\begin{table*}[!t]
    \centering
    \caption{Description of the ``Juntendo'' EEG dataset (Pt.: Patient ID; Age y.: Year, m.:Month; hr.: Hours, sec.: Second)}
    \label{tab:juntendo-data-table}
    \scalebox{0.78}
    {
    \begin{tabular}{c|c|c|c|c|c|r|r}
    \hline
        \textbf{Pt.} & 
        \textbf{Gender} & 
        \begin{tabular}[c]{@{}c@{}} \textbf{Age}\\\textbf{(y. m.)}\end{tabular} & 
        \begin{tabular}[c]{@{}c@{}} \textbf{Seizure} \\ \textbf{types} \end{tabular} & 
        \begin{tabular}[c]{@{}c@{}} \textbf{Channels with} \\ \textbf{seizure} \end{tabular} &
        \begin{tabular}[c]{@{}c@{}} \textbf{Record} \\ \textbf{length (hr.)} \end{tabular} & \begin{tabular}[c]{@{}c@{}}\textbf{Seizure}\\\textbf{time (sec.)}\end{tabular} & \begin{tabular}[c]{@{}c@{}}\textbf{Seizure} \\\textbf{ratio (\%)}\end{tabular} \\ \hline
         
         Pt-01 & Female &  19y 10m &\begin{tabular}[c]{@{}c@{}} Generalized,\\ Motor, Tonic-clonic \end{tabular} & 
         \begin{tabular}[c]{@{}c@{}}All Channels \\except Fp1,F7 \end{tabular} &  
         4 &   217 &  1.507 \\ \hline
         
         Pt-02 & Male &  16y 5m & \begin{tabular}[c]{@{}c@{}} Generalized, Motor,\\ Epileptic spasms\end{tabular} & 
         C3, F3, T3 &     
         4 &   247 &  1.715 \\ \hline
         
         Pt-03 & Male &  26y 11m & \begin{tabular}[c]{@{}c@{}}Focal, Nonmotor,\\Behavior arrest \end{tabular} & 
         F7, T3 &     
         8 &   371 &  1.288 \\ \hline
         
         Pt-04 & Male &   1y 4m & \begin{tabular}[c]{@{}c@{}} Generalized, Motor,\\ Epileptic spasms \end{tabular} & 
         All Channels &     
         2 &    17 &  0.236 \\ \hline

         Pt-05 & Male &   8y 5m & \begin{tabular}[c]{@{}c@{}}Focal, Nonmotor, \\ Behavior arrest \end{tabular}& 
         C3, F3 &     
         6 &   441 &  2.042 \\ \hline
         
         Pt-06 & Male &  30y 9m & \begin{tabular}[c]{@{}c@{}}Focal, Nonmotor,\\ Behavior arrest \end{tabular} & 
         F7, T3 &     
         6 &   416 &  1.926 \\ \hline

         Pt-07 & Male &  27y 0m & \begin{tabular}[c]{@{}c@{}}Focal, Nonmotor,\\Behavior arrest \end{tabular} & 
         F8, T4 &     
         4 &   161 &  1.118 \\ \hline
         
         Pt-08 & Male &  26y 4m & \begin{tabular}[c]{@{}c@{}} Focal, Motor,\\Automatisms \end{tabular}& 
         F7, T3 &     
         4 &   229 &  1.59 \\ \hline

         Pt-09 & Female &  19y 4m & \begin{tabular}[c]{@{}c@{}}Generalized, Motor,\\Tonic-clonic \end{tabular}& 
         \begin{tabular}[c]{@{}c@{}}All Channels \\except Fp1, F7 \end{tabular}&     
         2 &  1150 &  15.972 \\ \hline
         
        Pt-10 & Male &  24y 7m & \begin{tabular}[c]{@{}c@{}}Focal, Focal to \\bilateral tonic-clonic \end{tabular}& 
        F7, T3 &    
        10 &   218 &  0.606 \\ \hline
        
        Pt-11 & Female &  27y 8m & \begin{tabular}[c]{@{}c@{}}Focal, Motor,\\Automatisms\end{tabular}& 
        T5 &     
        4 &   343 &  2.382 \\ \hline
        
        Pt-12 & Male &  11y 7m & \begin{tabular}[c]{@{}c@{}}Focal, Motor, Tonic \end{tabular}& 
        \begin{tabular}[c]{@{}c@{}}F7,F8,\\T3,T4,T5,T6 \end{tabular}&  
        6.03 &  1581 &  7.282 \\ \hline

        Pt-13 & Female &  27y 9m & \begin{tabular}[c]{@{}c@{}}Focal, Motor,\\Automatisms \end{tabular}& 
        \begin{tabular}[c]{@{}c@{}}All Channels \\except Fz,Cz,Pz,\\C1,C2\end{tabular} &  
        2 &   135 &  1.875 \\ \hline

        Pt-14 & Male &  35y 5m & \begin{tabular}[c]{@{}c@{}}Focal, Motor,\\Automatisms \end{tabular}& 
        F7, F8 &    
        12 &   388 &  0.898 \\ \hline
        
        Pt-15 & Female &   5y 2m & \begin{tabular}[c]{@{}c@{}}Focal, Motor,\\Epileptic spasms \end{tabular}& 
        C4, F4 &     
        4 &   125 &  0.868 \\ \hline
        
        Pt-16 & Male &  23y 10m & \begin{tabular}[c]{@{}c@{}}Generalized, Motor,\\Tonic-clonic \end{tabular}& 
        All Channels &     
        2 &    94 &  1.306 \\ \hline
        
        Pt-17 & Male &  42y 0m & \begin{tabular}[c]{@{}c@{}}Generalized, Motor,\\Tonic-clonic \end{tabular}& 
        All Channels &     
        2 &    23 &  0.319 \\ \hline

        Pt-18 & Female &  12y 11m & \begin{tabular}[c]{@{}c@{}}Unknown, Motor,\\Epileptic spasms \end{tabular}& 
        C3, F3, F7, T3 &     
        4 &    59 &  0.41 \\ \hline
        
        Pt-19 & Female &  11y 8m & \begin{tabular}[c]{@{}c@{}}Generalized, Nonmotor,\\Typical absence \end{tabular}& 
        All Channels &     
        10 &   370 &  1.027 \\ \hline
        
        Pt-20 & Female &  35y 0m & \begin{tabular}[c]{@{}c@{}}Focal, Motor,\\Automatisms \end{tabular}& 
        T3, T5 &    
        10 &   495 &  1.375 \\ \hline
        
        Pt-21 & Male &  31y 6m & \begin{tabular}[c]{@{}c@{}}Focal, Motor, Tonic \end{tabular}& 
        T3, T4, T5, T6 &     
        2 &    27 &  0.375 \\ \hline
        
    \end{tabular}}
\end{table*}

\subsection{Preprocessing}
\label{sec:preprocessing}
As a preprocessing step, each EEG recording was filtered with an FIR bandpass filter from 1 to 60 Hz. In addition, a 50-Hz notch filter was applied to eliminate power noise from the recording system. The filtered EEG recordings were segmented with a fixed-length window (non-overlapping), with the window length depending on the experiment. In total, 19 commonly used monopolar channels were selected for the experiment (see Section \ref{sec:siena_dataset}).

\section{Proposed Model}
\label{sec:proposed_model}

\subsection{MR-EEGWaveNet: Multiresolutional EEGWaveNet}
\label{sec:multireso_EEGWaveNet}

\begin{table*}[t]
    \centering
    \caption{MR-EEGWaveNet architecture}
    \label{tab:multireso_EEGWaveNet_arch}
    \scalebox{0.8}{
    \begin{tabular}{cccccc}
    \hline
      \textbf{Module} 
      &  \textbf{Layer} 
      &  \textbf{Kernel} 
      &  \textbf{Output} 
      &  \textbf{Activation} \\
    \hline
          \textbf{A}  &        \textbf{Convolution Module}        &    &                  &                       \\\hline
          \textit{I)} &    \textit{Multi-scale Convolution}       &    &                  &                       \\
                      &           \textit{Input} &                              &           ($C$, $N$) &               Linear  \\
                      &          Conv1D &  kernel 2, stride 2, group $C$ &         ($C$, $N$/2) &               Linear  \\
                      &          Conv1D &  kernel 2, stride 2, group $C$ &         ($C$, $N$/4) &               Linear  \\
                      &          Conv1D &  kernel 2, stride 2, group $C$ &         ($C$, $N$/8) &               Linear  \\
                      &          Conv1D &  kernel 2, stride 2, group $C$ &         ($C$, $N$/16)&               Linear  \\
                      &          Conv1D &  kernel 2, stride 2, group $C$ &         ($C$, $N$/32)&               Linear  \\
                      &          Conv1D &  kernel 2, stride 2, group $C$ &         ($C$, $N$/64)&               Linear  \\
    \hline
                     \textit{II)} & \textit{Spatio-temporal Convolution} & & &\\
                      &  \textit{Input} &                          &   ($C$, $N$/$2^k$)   &                      \\
                      &          Conv1D &        (32$\times$4) &  (32, $N$/$2^k-3$) &                      \\
                      &     BatchNorm1D &                          &                   &                      \\
                      &      Activation &                          &                   &           LeakyReLU(0.01) \\
                      &          Conv1D &        (32$\times$4) &  (32, $N$/$2^k-6$) &                      \\
                      &     BatchNorm1D &                          &                   &                      \\
                      &      Activation &                          &                   &           LeakyReLU(0.01) \\
                      & Global average pooling &                   &                32 &                      \\
    \hline

           \textbf{B} & \textbf{Feature Extraction Module} &       &                   &                      \\\hline
                      &  \textit{Input} &                          &               160 &                      \\
                      & Fully Connected &                          &                64 &           LeakyReLU(0.01) \\
                      & Fully Connected &                          &                32 &               \\
                      & Normalization   &                          &                32 &               \\\hline
                      
           \textbf{C} & \textbf{Predictor Module} &                &                   &                      \\\hline
                      &           \textit{Input} &                 &                 K &                      \\
                      & Fully Connected &                          &                64 &           LeakyReLU(0.01)  \\
                      & Fully Connected &                          &                32 &              Sigmoid \\
                      & Classifier      &                          &                 2 &          Log Softmax \\
    \hline
    \end{tabular}
    }
\end{table*}

\begin{figure}[th!]
    \centering
    \begin{subfigure}{1.0\textwidth}
        \centering
        \centerline{\includegraphics[width=1.0\linewidth]{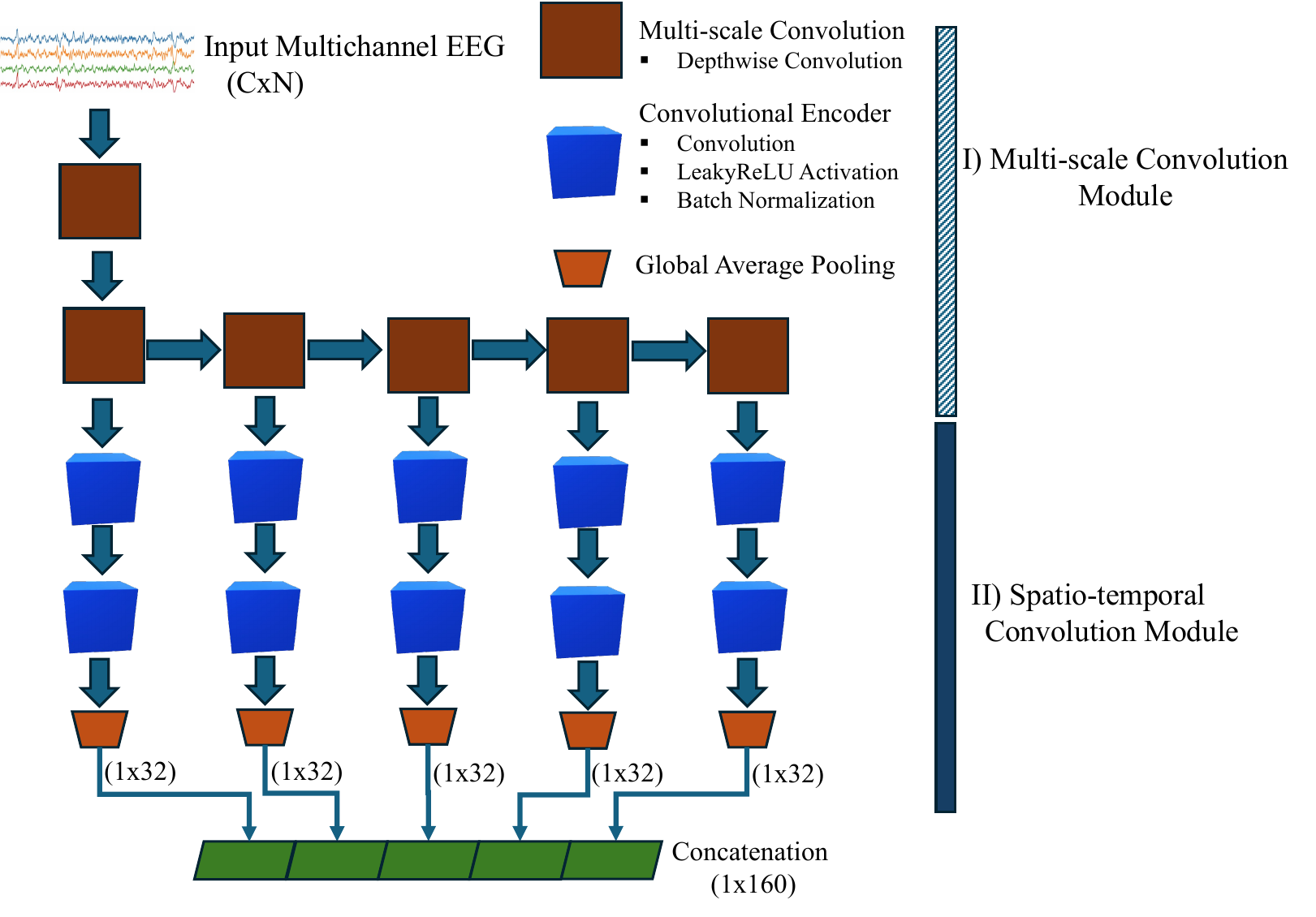}}
        \caption{Convolution module}
        \label{fig:conv_module}
    \end{subfigure}
    
    \vspace{1em}
    
    \begin{subfigure}{1.0\textwidth}
        \centering
        \centerline{\includegraphics[width=1.0\linewidth]{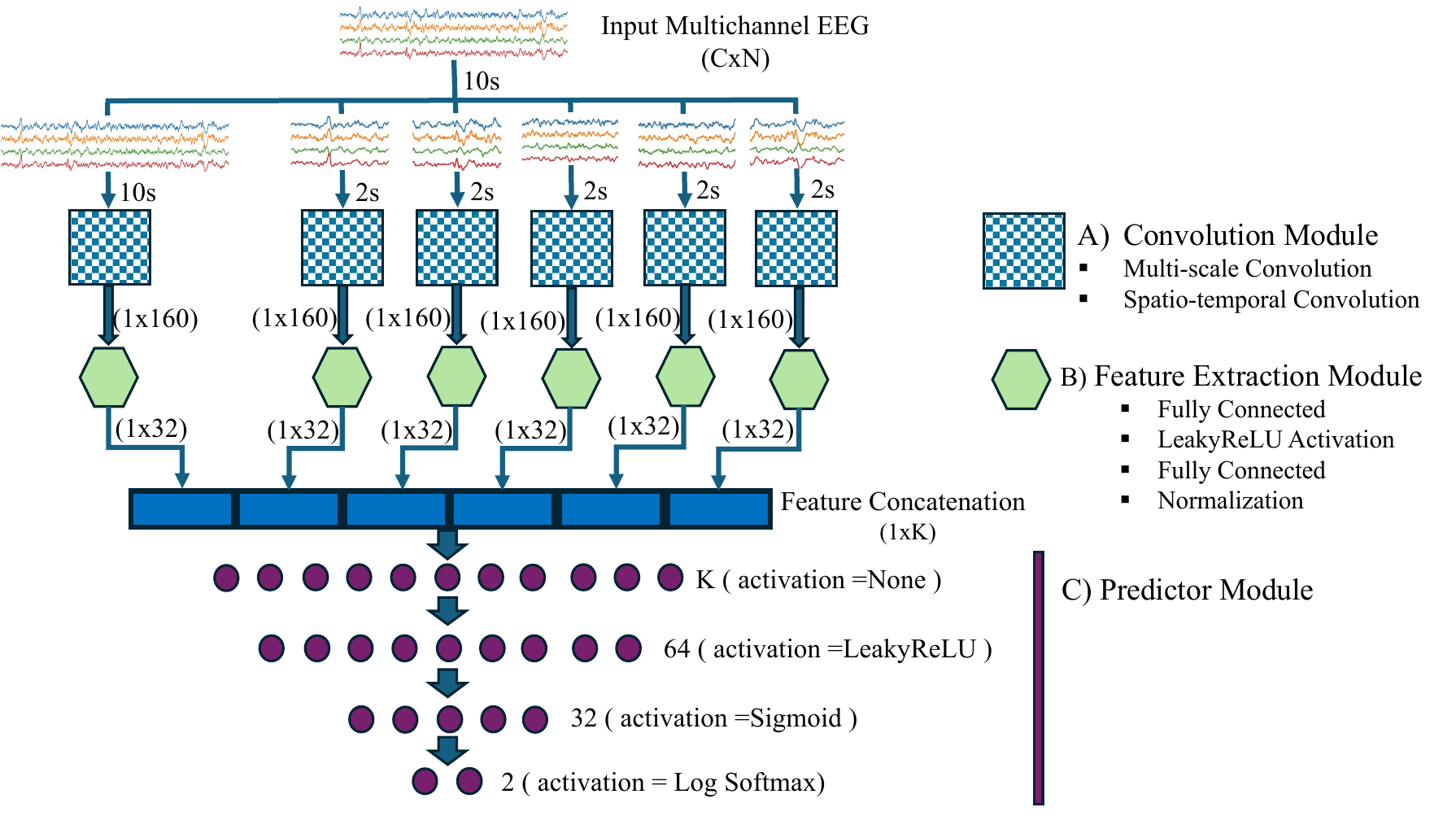}}
        \caption{MR-EEGWaveNet architecture}
        \label{fig:mr-eegwavenet-arch-image}
    \end{subfigure}

\caption{Classification process of MR-EEGWaveNet model (a) Convolution module, and (b) Overall MR-EEGWaveNet architecture (Here, $10$ s, and $2$ s represent $10$-second, and $2$-second long multichannel EEG signal)}
\label{fig:mr-eegwavenet-arch}
\end{figure}

The architecture of MR-EEGWaveNet is shown in Table \ref{tab:multireso_EEGWaveNet_arch}.
The MR-EEGWaveNet model contains three main modules: convolution, feature extraction, and predictor. The ``convolution'' module has a series of convolutional layers and pooling operations to process EEG signals, and is divided into two parts: a multi-scale convolution module and a spatio-temporal convolution module. The ``multi-scale convolution'' module has six temporal convolutional layers, each reducing the signal sequence length while maintaining channel information. The output of each temporal convolutional layer feeds to the next convolutional layer (except the last convolutional layer). The ``spatio-temporal convolution'' module performs pooling operations across multiple blocks to extract features from all channels. In each block, there are three operational components: convolutional layers, batch normalization, and LeakyReLU activation. For the MR-EEGWaveNet architecture, we used two blocks sequentially to calculate features. Here, the inputs are the outputs of the temporal convolutional layers from ``multi-scale convolution'' (except the first temporal convolutional layer). Thus, five ``spatio-temporal convolution'' modules are used in the proposed model, each producing an output of $1 \times 32$. Subsequently, they are concatenated to form a vector of dimension $1 \times 160$. 

Thereafter, the ``feature extraction'' module shrank the dimension of the feature from $160$ to $32$ with two fully connected layers, followed by normalization. Thus, the final output of modules A and B for an EEG segment is $32$. The same process applies to the sub-segments. For example, if the proposed model has multiresolutional parameters $10$ s and $2$ s, then it will calculate one feature vector ($1 \times 32$) for $10$ s and five feature vectors corresponding to five sub-segments of $2$ s (each has the same dimension of $1 \times 32$). After concatenation, the final size of the feature vector will be $1 \times 192$. Here, the list of multiresolutional parameters defines the architecture of the model. Figure \ref{fig:mr-eegwavenet-arch} shows the architecture for the parameter list ``$10$ s, $2$ s''. In this case, the model subdivides the $10$-s EEG signal into $2$ s segments; in total, six EEG signals (one $10$ s and five $2$ s multichannel EEG signals) are generated. Each signal is fed to modules A through B to calculate the features of each of them. Finally, the concatenated feature vector is fed into the predictor module for classification. The predictor module contains a fully connected classifier that maps the features to the output classes. The class probabilities are computed using the log-softmax function. The convolution module and the detailed architecture of the MR-EEGWaveNet for the above example are shown in Figures \ref{fig:conv_module} and \ref{fig:mr-eegwavenet-arch-image}, respectively. The main difference between EEGWaveNet and MR-EEGWaveNet architecture is that MR-EEGWaveNet extracts features both from the EEG segments and their sub-segments using modules A and B (the convolution and feature extraction module). Thus, the EEG segment is segmented into multiple levels based on the parameters. 

\subsection{Post-Classification Processing}
\label{sec:post_processing}
To further improve the performance of the proposed architecture, we applied an anomaly score-based post-classification processing technique to reduce the false detection rate of seizures. Therefore, we calculated the anomaly scores of the EEG segments, $a_i$ ($i$ is the segment index), using one of the state-of-the-art anomaly detection methods, ECOD: Unsupervised Outlier Detection Using Empirical Cumulative Distribution Functions \cite{ECOD_Li_2022}. Finally, the following rule is applied to decide whether a segment is classified as a seizure or nonseizure based on the anomaly score of the EEG segments:
\begin{align}
    \mu_{a} &= \frac{1}{M} \sum_{i=1}^{M} a_{i} \label{eq:anomaly_mean}\\ 
    L_{i} &= 
    \begin{cases}
        1, & \text{seizure, if } a_{i}> \mu_{a}  \\ 
        0, & \text{nonseizure}.
    \end{cases}\label{eq:anomaly_th}
\end{align}
where $\mu_{a}$ is the mean of $a_{i}$, $M$ is the total number of segments, and $L_{i}$ is the predicted label of a segment.

\section{Experiment}
\label{sec:experiment}
The raw EEG data from each patient were preprocessed and segmented into equal length, $W$ (window length in seconds), using a sliding window approach with no overlap. Each segment of EEG has a dimension of $C \times N$, where $C$ is the number of channels, and $N$ is the number of time points of a single channel. In the MR-EEGWaveNet (see Table \ref{tab:multireso_EEGWaveNet_arch}), $C=19$, and $k$ is the order of layers in Module A(II). Let $F_{s}$ be the sampling frequency, then $N = WF_{s}$.
 
For EEGWaveNet, the feature dimension is $F$=$32$, whereas in the case of MR-EEGWaveNet, it depends on the multiresolution parameter list $D$ (such as [$10$ s, $5$ s, $2$ s]). The number of parameters and the computational complexity of the models depend on $D$, $F$, and $W$. Here, the concatenated feature length $K$ depends on them and can be calculated as:
\begin{align} \label{eq:feat_len_cal}
    K = \sum_{d\in D}F \left\lfloor\frac{W}{d} \right\rfloor 
\end{align}
where $\left\lfloor \cdot \right\rfloor$ denotes the floor function producing the largest integer less than or equal to the argument.

Six different models were evaluated on the datasets. For EEGWaveNet we have experimented with  EEGWaveNet-1 {[}2 s{]}, EEGWaveNet-2 {[}5 s{]}, and EEGWaveNet-3 {[}10 s{]} models corresponding to window lengths of $2$ s, $5$ s, and $10$ s, respectively. In the case of the proposed model, the three experimented models are listed as follows: MR-EEGWaveNet-1 {[}5 s, 2.5 s{]}, MR-EEGWaveNet-2 {[}10 s,  2 s{]}, and MR-EEGWaveNet-3 {[}10 s, 5 s, 2 s{]}. 

Moreover, to evaluate the contribution of individual components within the MR-EEGWaveNet architecture, we conducted an ablation study using two simplified variants: MR-EEGWaveNet-Abl1 and MR-EEGWaveNet-Abl2. We selected the MR-EEGWaveNet-2 model as the baseline, which guided both the development of ablation variants and the experimental comparison. The two ablation models are defined as follows:
\begin{enumerate}[label=\roman*)]
    \item \textbf{MR-EEGWaveNet-Abl1} excludes the feature extraction module from the baseline model.
    \item \textbf{MR-EEGWaveNet-Abl2} removes the $10$-second input pipeline, retaining only the $2$-second sub-segments stream.
\end{enumerate}
The detailed architectures of the ablated models, MR-EEGWaveNet-Abl1 and MR-EEGWaveNet-Abl2, are provided in Appendix~\ref{appen:ablation_model}, in Figures~\ref{fig:mr-eegwavenet-abl1-arch-image} and~\ref{fig:mr-eegwavenet-abl2-arch-image}, respectively.

In the case of calculating anomaly scores using ECOD, we applied it directly to the preprocessed EEG segments rather than to learned embeddings. For each segment, all channels were concatenated into a single feature vector by flattening the multichannel signal ($C$ channels × $N$ time samples per channel $\rightarrow$ $C \times N$-dimensional vector). The flattened signals were z-score normalized (standardized to have zero mean and unit variance) before further processing. ECOD was then applied to these fixed-length vectors to compute anomaly scores for each segment in a recording, which were classified as anomalous based on Equations \ref{eq:anomaly_mean} and \ref{eq:anomaly_th}. The ECOD implementation can be found at \cite{zhao2024pyod2, zhao2019pyod}.

The performance of the models was evaluated with the metrics: precision (Pre.), recall (Rec.), specificity (Spe.), F1 score (F1), false positive rate (FPR), seizure detection ratio (Det. Ratio), and receiver operating characteristic--area under the curve (AUC).
\begin{align}
\mathrm{Precision} &= \frac{TP}{TP+FP}\\
\mathrm{True\ Positive\ Rate\ (Recall)} &= \frac{TP}{TP+FN}\\
\mathrm{Specificity} &= \frac{TN}{TN+FP}\\
\mathrm{F1\ Score} &= 2 \times \frac{\mathrm{Precision} \times \mathrm{Recall}}{\mathrm{Precision}+\mathrm{Recall}}\\
\mathrm{False\ Positive\ Rate\ (FPR)} &= \frac{FP}{FP + TN}\\
\text{Det. Ratio} &=  \frac{\text{Number of detected seizures}}{\text{Total number of seizures}} 
\end{align}
Where $TP$, $TN$, $FP$, and $FN$ denote the total number of true positives, true negatives, false positives, and false negatives, respectively, based on the detection of EEG segments. Moreover, the Receiver Operating Characteristic--Area Under the Curve (ROC-AUC) is a standard metric used to evaluate the ability of a binary classification model to differentiate between two classes.   
The ROC curve plots the True Positive Rate (TPR) (also known as recall) against the FPR at various classification thresholds. The Area Under the Curve (AUC) measures the entire area beneath the ROC curve and quantifies the overall performance of a model. The AUC is a single value that ranges from 0 to 1. An AUC of 1.0 indicates perfect classification, while an AUC of 0.5 does not reflect discriminative ability, the same as random guessing. Generally, a classifier with a higher AUC is considered to be more effective in binary classification tasks. Finally, the $\text{Det. Ratio}$ represents the seizure detection ratio from the outcome of a model.

\begin{figure}[!t]
    \centering
    \includegraphics[width=0.65\linewidth]{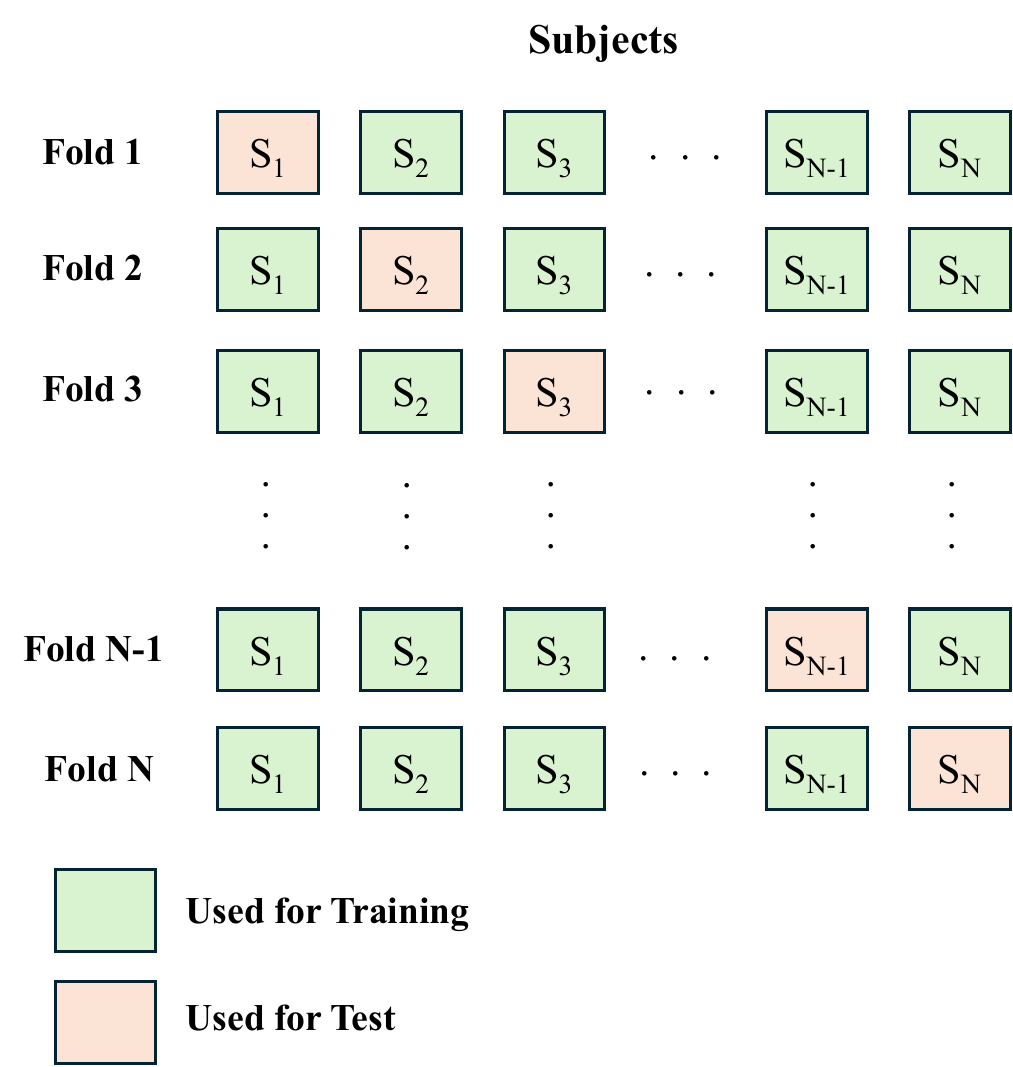}  
    \caption{Illustration of Leave-One-Subject-Out (LOSO) cross-validation scheme}
    \label{fig:LOSO}
\end{figure}

The models were evaluated with a LOSO validation approach, where in each fold, one patient was designated for testing and the remaining patients were used for training, as illustrated in Figure~\ref{fig:LOSO}.
In the case of preparing a dataset for training, we adopted the following two techniques to overcome the data imbalance problem: 
\begin{enumerate}[label=\Roman*.]
    \item \textbf{Seizure data:} Oversampling; the seizure segments (within the expert's notation of the seizure label) are $80\%$-overlapped between consecutive segments.   
    \item \textbf{Nonseizure data:} Undersampling; the nonseizure segments (non-overlapping) are randomly selected from interictal EEG \cite{interictal_EEG_2010}, the ratio between seizure and nonseizure is 1:2.
\end{enumerate}
For the test set, the recordings were segmented using a non-overlapping sliding window approach with a fixed window size $W$, where each segment was assigned a label either $0$ (nonseizure, negative class) or $1$ (seizure, positive class) based on the ground truth label provided by an expert (similar to \cite{Sigsgaard2023_SZ_detection}).  

We used the following parameter values to train each model: batch size: $32$, number of training epochs: $300$, and Adam optimizer with learning rate: $10^{-3}$. The cross-entropy loss function was employed with balanced weights of $0.75$ and $1.5$ for the nonseizure and seizure classes, respectively. The optimizer was reinitialized at the beginning of each training epoch, effectively resetting the optimization state while retaining the current model parameters. An early stopping criterion was employed, where training was stopped if the validation loss did not improve for $30$ consecutive epochs (patience = $30$). For each training epoch, nonseizure samples were randomly selected at twice the number of seizure samples (i.e., a 1:2 seizure-to-nonseizure ratio). This balanced subset is then randomly split: $70\%$ for training and $30\%$ for validation. Each experiment was repeated three times to ensure stability of the results. For each patient, the performance metrics were averaged across the three runs; that is, the final score for a patient was computed as the mean of their results over the three iterations.

The state-of-the-art CBraMod foundation model \cite{wang2025cbramod} was included in the evaluation for performance comparison with the proposed MR-EEGWaveNet.
We employed the pretrained CBraMod model, trained on the large public Temple University Hospital EEG corpus (TUEG) \cite{TUSZ_2016}, and used it in a frozen state to extract features, without performing any fine-tuning on the target datasets. The TUEG dataset consists of a diverse collection of clinical EEG recordings from 14,987 subjects, totaling 27,062 hours in duration.
As preprocessing, 19 common EEG channels (Fp1, Fp2, F7, F3, Fz, F4, F8, T3, C3, Cz, C4, T4, T5, P3, Pz, P4, T6, O1, and O2) were chosen in accordance with the 10-20 electrode placement standards to ensure clean and standardized pre-training input. Power line noise was removed using a $60$ Hz notch filter, and low/high-frequency noise was suppressed using a band-pass filter ($0.3$--$75$ Hz). EEG signals were resampled to $200$~Hz, with segments exceeding $\pm100~\mu$V removed for clean pre-training. Signals were then normalized to $\pm1$ by setting $100~\mu$V as the unit. The length of each EEG patch was set to $1$ second (corresponding to $200$ data points). The detailed pre-training and preprocessing procedures can be found in \cite{wang2025cbramod}. 

We also followed the similar preprocessing strategy on the experimental dataset to extract features using the CBraMod model, with the exception of using $50$-Hz notch filtering and not removing segments exceeding $\pm100~\mu$V as in pre-training. Later, the features are classified with the Light Gradient Boosting Machine (LightGBM) \cite{lightGBM_2017_Ke_Guolin} classifier. The LightGBM classifier is a free, open-source gradient boosting framework. It is particularly appropriate for classification or regression tasks on large-scale, high-dimensional datasets. The LightGBM classifier was evaluated using the LOSO approach, with the training data constructed at a seizure:nonseizure ratio of 1:2 (classes 1 and 0), as previously described. The detailed LightGBM parameter list is provided in Appendix \ref{appen:lightGBM} (Table \ref{tab:lightGBM_default_param}).

We implemented the experiments based on the Python 3.9.13 and PyTorch 2.7.1+CUDA 12.6. The models were trained on the following machines: 
(I) an Intel Xeon w7-3565X 2.5$\sim$4.8GHz (RAM 256GB) CPU with two NVIDIA RTX A6000 Ada (48GB) GPUs, 
(II) 2x Intel(R) Xeon(R) Silver 4310 CPU @ 2.10GHz (RAM 64GB) with four NVIDIA RTX A6000 (48GB) GPUs, 
(III) an AMD EPYC 7742 2.25$\sim$3.4 GHz 64-core (RAM 512GB) CPU with four NVIDIA A100 (80GB) GPUs, and
(IV) an Intel Xeon W-2195 CPU @ 2.30$\sim$4.3GHz (RAM 256GB) with two NVIDIA TITAN RTX (24GB)GPUs.

\section{Results}
\label{sec:results}

\subsection{Experimental Objectives}
\label{sec:res_exp_objectives}
The objective of our experiments was to investigate the impact of multiresolution parameters on the performance of the classification model. We considered the following aspects when selecting parameters for the experiments:
 \begin{enumerate}[label=\roman*)] 
     \item \textbf{Effect of Window Length on Model Performance}: The model behavior with different window lengths\label{item:aspect1}.
     \item \textbf{Performance of the Proposed Model}: Achieving better performance using the proposed model.
     \item \textbf{Contribution of Post-Classification Processing}: Contributions made by the post-classification processing method to the models.
 \end{enumerate}

\subsection{Effect of Window Length on Model Performance}
\label{sec:res_window_length_effect}
To investigate the aspect \ref{item:aspect1}, we considered the window lengths of $2$ s, $5$ s and $10$ s as parameters in the EEGWaveNet model, and their corresponding models were EEGWaveNet-1 {[}2 s{]}, EEGWaveNet-2 {[}5 s{]}, and EEGWaveNet-3 {[}10 s{]}, respectively.
In the Siena dataset (see Table \ref{tab:comp_perf_siena_wo_preprocess}), performance improves consistently with longer window lengths. Among all, EEGWaveNet-3 achieves the best results with the $10$-s window, indicating the contribution of the longer window size to the seizure detection task. A similar improvement in performance (except for specificity) with longer window lengths can be observed in the Juntendo dataset (see Table \ref{tab:comp_perf_juntendo_wo_preprocess}). In this case, the specificity decreases by approximately 3\% and 2.6\% when the window length increases from 2 to 5 s and from 2 to 10 s, respectively. 

Furthermore, we investigated our MR-EEGWaveNet with two different window length sizes and three different multiresolutional parameters, which are as follows: MR-EEGWaveNet-1 {[}5 s, 2.5 s{]}, MR-EEGWaveNet-2 {[}10 s,  2 s{]}, and MR-EEGWaveNet-3 {[}10 s, 5 s,  2 s{]}. The window length of model MR-EEGWaveNet-1 was $5$ s, where both MR-EEGWaveNet-2 and MR-EEGWaveNet-3 were $10$ s long. For the Siena dataset (see Table \ref{tab:comp_perf_siena_wo_preprocess}), all metrics except recall improved with longer window lengths, while recall decreased by approximately 2\% in the MR-EEGWaveNet-2 and MR-EEGWaveNet-3 models compared to MR-EEGWaveNet-1. For the Juntendo dataset (see Table \ref{tab:comp_perf_juntendo_wo_preprocess}), similar improvements in performance were observed with longer window lengths.

\begin{table}[t!]
\centering
\caption{Performance comparison between EEGWaveNet and the MR-EEGWaveNet on the Siena dataset.}
\label{tab:comp_perf_siena_wo_preprocess}
\scalebox{1.0}
{
\begin{tabular}{lccccc}
\hline
      \textbf{Models [Segments]}
    & \textbf{Pre.} 
    & \textbf{Rec.}	
    & \textbf{Spe.}	
    & \textbf{F1}
    & \textbf{AUC}
    \\\hline
    EEGWaveNet-1 {[}2 s{]}  	        & 5.82      & 77.92	   & 83.63   & 0.098  & 0.8614  \\
    EEGWaveNet-2 {[}5 s{]}  	        & 8.05      & 84.84    & 84.25   & 0.126  & 0.9097  \\
    EEGWaveNet-3 {[}10 s{]}             & 11.72     & 85.99    & 86.04   & 0.161  & 0.9188  \\\hline\hline

    MR-EEGWaveNet-1 {[}5 s, 2.5 s{]}      & 15.58     & 84.02    & 90.43   & 0.218  & 0.9291 \\
    MR-EEGWaveNet-2 {[}10 s,  2 s{]}  	    & 26.74     & 82.39    & 93.86   & 0.319  & 0.9421 \\
    MR-EEGWaveNet-3 {[}10 s, 5 s,  2 s{]}   & 25.48     & 82.06    & 93.15   & 0.316  & 0.9441 \\\hline 
\end{tabular}
}
\end{table}

\begin{table}[t!]
\centering
\caption{Performance comparison of EEGWaveNet and MR-EEGWaveNet on the Juntendo Dataset.}
\label{tab:comp_perf_juntendo_wo_preprocess}
\scalebox{1.0}
{
\begin{tabular}{lccccc}
\hline
      \textbf{Models [Segments]}
    & \textbf{Pre.} 
    & \textbf{Rec.}	
    & \textbf{Spe.}	
    & \textbf{F1}
    & \textbf{AUC}
    \\\hline

    EEGWaveNet-1 {[}2 s{]}  	    & 18.20  & 79.82 & 85.93 & 0.253 & 0.8823  \\
    EEGWaveNet-2 {[}5 s{]}       & 18.51  & 82.04 & 83.03 & 0.254 & 0.8962  \\
    EEGWaveNet-3 {[}10 s{]}  	& 19.07  & 88.83 & 83.35 & 0.274 & 0.9266  \\\hline
    \hline
    MR-EEGWaveNet-1 {[}5 s, 2.5 s{]}    & 36.13  & 76.91 & 91.99 & 0.428 & 0.9279 \\
    MR-EEGWaveNet-2 {[}10 s,  2 s{]}  	  & 42.92  & 77.94 & 91.60 & 0.474 & 0.9414 \\ 
    MR-EEGWaveNet-3 {[}10 s, 5 s,  2 s{]} & 43.75  & 78.60 & 92.61 & 0.488 & 0.9410 \\\hline
\end{tabular}
}
\end{table}

\subsection{Performance of the Proposed Model}
\label{sec:res_perf_proposed_method}
A comparative analysis between EEGWaveNet and MR-EEGWaveNet is presented in Tables \ref{tab:comp_perf_siena_wo_preprocess} and \ref{tab:comp_perf_juntendo_wo_preprocess} for the Siena dataset and the Juntendo dataset, respectively. 
As shown in Table \ref{tab:comp_perf_siena_wo_preprocess}, the proposed MR-EEGWaveNet achieves better performance than the baseline EEGWaveNet on the Siena dataset. While EEGWaveNet improves slightly with longer segment lengths, MR-EEGWaveNet has considerably higher precision (up to 26.74\%) and specificity (up to 93.86\%). Although the recall of MR-EEGWaveNet is slightly lower than EEGWaveNet-3 (82.39\% vs. 85.99\%), it achieves higher F1 scores (up to 0.319) and AUC values (up to 0.9441). 
Similarly, as presented in Table \ref{tab:comp_perf_juntendo_wo_preprocess}, MR-EEGWaveNet shows better performance over EEGWaveNet on the Juntendo dataset. In this case, MR-EEGWaveNet outperforms EEGWaveNet with significantly higher precision (up to 43.75\%), specificity (up to 92.61\%), F1 scores (up to 0.488), and AUC values (up to 0.9414). However, MR-EEGWaveNet shows lower recall than EEGWaveNet but achieves a better balance with higher specificity (92.61\%) compared to EEGWaveNet's 83.35\%, which is considerably lower.

\begin{table}[t!]
\centering
\caption{Performance comparison between EEGWaveNet and the MR-EEGWaveNet, with and without the ``post-classification processing'' technique, on the Siena dataset.}
\label{tab:comp_perf_siena_all}
\scalebox{1.0}
{
\begin{tabular}{lccccc}
\hline
      \textbf{Models [Segments]}
    & \textbf{Pre.} 
    & \textbf{Rec.}	
    & \textbf{Spe.}	
    & \textbf{F1}
    & \textbf{AUC}
    \\\hline
    EEGWaveNet-1 {[} 2 s{]}  	                     & 5.82      & 77.92	& 83.63   & 0.098  & 0.8614  \\
    EEGWaveNet-1 {[} 2 s{]} (post-process)	     & 7.21      & 65.55	& 90.55   & 0.118  & --  \\\hline
    
    EEGWaveNet-2 {[}5 s{]}  	                     & 8.05      & 84.84    & 84.25   & 0.126  & 0.9097  \\
    EEGWaveNet-2 {[}5 s{]} (post-process) 	     & 9.59      & 81.71    & 89.64   & 0.149  & --  \\\hline
    
    EEGWaveNet-3 {[}10 s{]}                       & 11.72     & 85.99    & 86.04   & 0.161  & 0.9188  \\
    EEGWaveNet-3 {[}10 s{]} (post-process)        & 12.80     & 82.35    & 89.67   & 0.177  & --  \\
    \hline
    \hline

    MR-EEGWaveNet-1 {[}5 s, 2.5 s{]}                    & 15.58     & 84.02    & 90.43   & 0.218  & 0.9291 \\
    MR-EEGWaveNet-1 {[}5 s, 2.5 s{]} (post-process)     & 18.93     & 81.16    & 92.89   & 0.252  & -- \\\hline
    
    MR-EEGWaveNet-2 {[}10 s,  2 s{]}  	                 & 26.74     & 82.39    & 93.86   & 0.319  & 0.9421 \\
    MR-EEGWaveNet-2 {[}10 s,  2 s{]} (post-process)      & 28.70     & 79.66    & 95.05   & 0.336  & -- \\\hline
    
    MR-EEGWaveNet-3 {[}10 s, 5 s,  2 s{]}                 & 25.48     & 82.06    & 93.15   & 0.316  & 0.9441 \\
    MR-EEGWaveNet-3 {[}10 s, 5 s,  2 s{]} (post-process)  & 27.48     & 79.49    & 94.92   & 0.335  & -- \\\hline
    
\end{tabular}
}
\end{table}

\begin{table}[t!]
\centering
\caption{Performance comparison between EEGWaveNet and the MR-EEGWaveNet, with and without the ``post-classification processing'' technique, on the Juntendo dataset.}
\label{tab:comp_perf_juntendo_all}
\scalebox{1.0}
{
\begin{tabular}{lccccc}
\hline
      \textbf{Models [Segments]}
    & \textbf{Pre.} 
    & \textbf{Rec.}	
    & \textbf{Spe.}	
    & \textbf{F1}
    & \textbf{AUC}
    \\\hline

    EEGWaveNet-1 {[} 2 s{]}  	                      & 18.20  & 79.82 & 85.93 & 0.253 & 0.8823  \\
    EEGWaveNet-1 {[} 2 s{]} (post-process) 	      & 23.69  & 78.68 & 92.47 & 0.315 & --  \\\hline
    
    EEGWaveNet-2 {[}5 s{]}                         & 18.51  & 82.04 & 83.03 & 0.254 & 0.8962  \\
    EEGWaveNet-2 {[}5 s{]} (post-process)          & 22.72  & 80.86 & 91.19 & 0.302 & --  \\\hline

    EEGWaveNet-3 {[}10 s{]}  	                  & 19.07  & 88.83 & 83.35 & 0.274 & 0.9266  \\
    EEGWaveNet-3 {[}10 s{]} (post-process) 	      & 23.56  & 87.91 & 90.94 & 0.327 & --  \\\hline
    \hline
    MR-EEGWaveNet-1 {[}5 s, 2.5 s{]}                     & 36.13  & 76.91 & 91.99 & 0.428 & 0.9279 \\
    MR-EEGWaveNet-1 {[}5 s, 2.5 s{]} (post-process)      & 39.46  & 76.64 & 94.76 & 0.455 & -- \\\hline
    
    MR-EEGWaveNet-2 {[}10 s,  2 s{]}  	                  & 42.92  & 77.94 & 91.60 & 0.474 & 0.9414 \\ 
    MR-EEGWaveNet-2 {[}10 s,  2 s{]} (post-process) 	  & 44.18  & 77.57 & 94.51 & 0.488 & -- \\\hline

    MR-EEGWaveNet-3 {[}10 s, 5 s,  2 s{]}                  & 43.75  & 78.60 & 92.61 & 0.488 & 0.9410 \\
    MR-EEGWaveNet-3 {[}10 s, 5 s,  2 s{]} (post-process)   & 46.07  & 78.26 & 95.40 & 0.509 & -- \\\hline

\end{tabular}
}
\end{table}

\subsection{Contribution of the Post-Classification Processing}
\label{sec:res-post-class-contribution}
\begin{figure}[t!]
    \centering
    \includegraphics[width=1.0\textwidth]{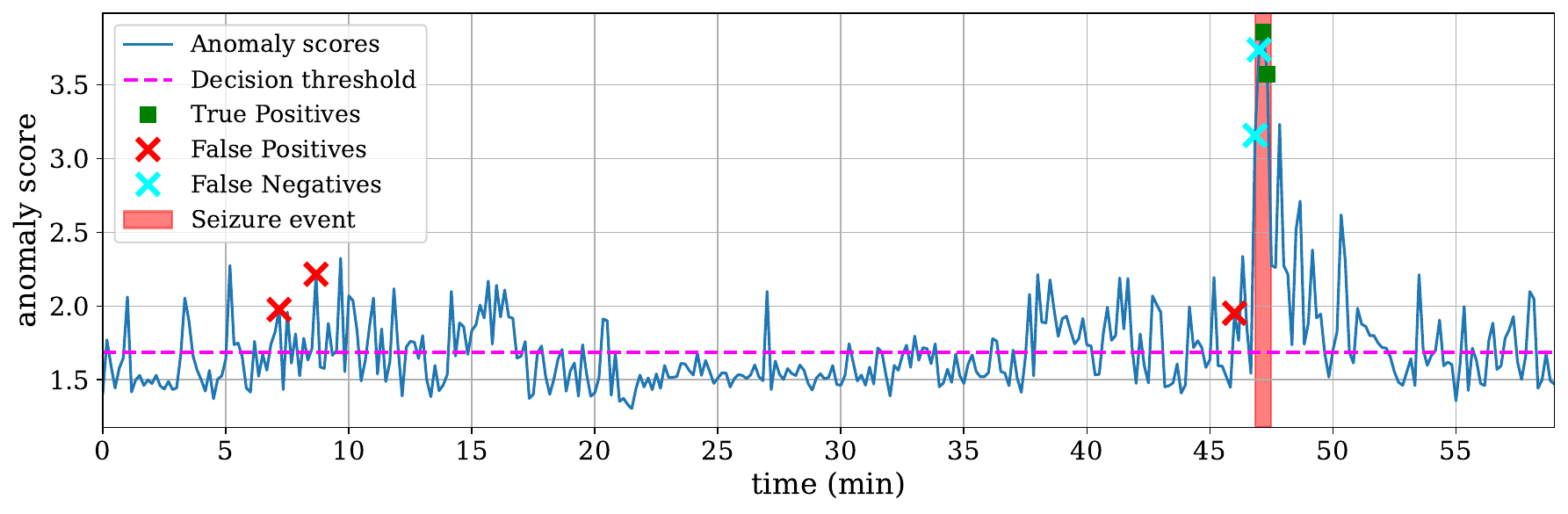}
    \caption{Outcome of the MR-EEGWaveNet-2 model on a 1-hour EEG recording after post-classification processing. \label{fig:mr_eegwavenet_pcp_outcome}}
\end{figure}

Figure \ref{fig:mr_eegwavenet_pcp_outcome} presents the final output of the MR-EEGWaveNet model on a 1-hour EEG recording following post-classification processing.
Table~\ref{tab:comp_perf_siena_all} shows a comparison between the baseline EEGWaveNet architecture and the MR-EEGWaveNet, both with and without post-classification processing, evaluated on the Siena dataset. In the case of proposed models, post-classification processing further enhances performance, particularly in terms of precision, specificity, and F1 score. Post-classification processing slightly affects recall, while specificity improves by approximately $2$--$3\%$.  Similarly to the proposed models, for all EEGWaveNet models (EEGWaveNet-1 to 3), the post-classification processing consistently improves precision and specificity, with a slight drop in recall $2$--$3\%$ (except for model EEGWaveNet-1 which is approximately $12\%$, a possible explanation is that shorter windows within the seizure event may yield low anomaly scores for certain segments). 

In Table~\ref{tab:comp_perf_juntendo_all}, we present a similar comparative study between EEGWaveNet and MR-EEGWaveNet in the Juntendo dataset. The experimental results demonstrated performance improvements across all metrics (except recall, which remains almost the same), similar to the Siena dataset. 
The post-classification processing improved the specificity of the proposed model by approximately $3\%$, and by $7-8\%$ for the EEGWaveNet models (although recall dropped slightly by about $1-2\%$). Those performance improvements indicate the effectiveness of the post-classification processing technique with the EEGWaveNet and the proposed models.

\subsection{Feature Visualization}
\label{sec:feat_visualization}

\begin{figure}[t!]
    \centering
    \begin{subfigure}{0.75\textwidth}
        \centering
        \includegraphics[width=\textwidth]{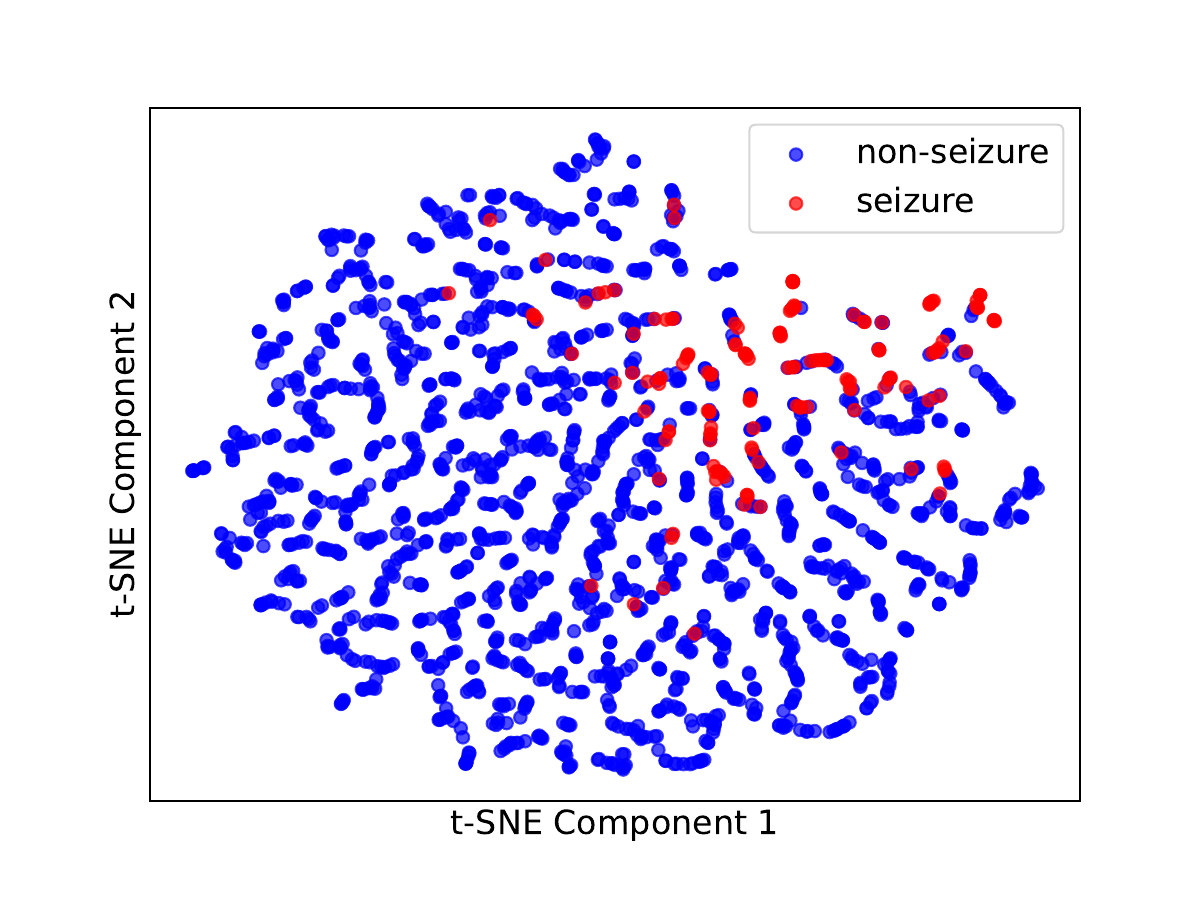}
        \caption{EEGWaveNet \label{fig:feat_vis_eegwavenet}}
    \end{subfigure}
    \vspace{0.1cm}
    \begin{subfigure}{0.75\textwidth}
        \centering
        \includegraphics[width=\textwidth]{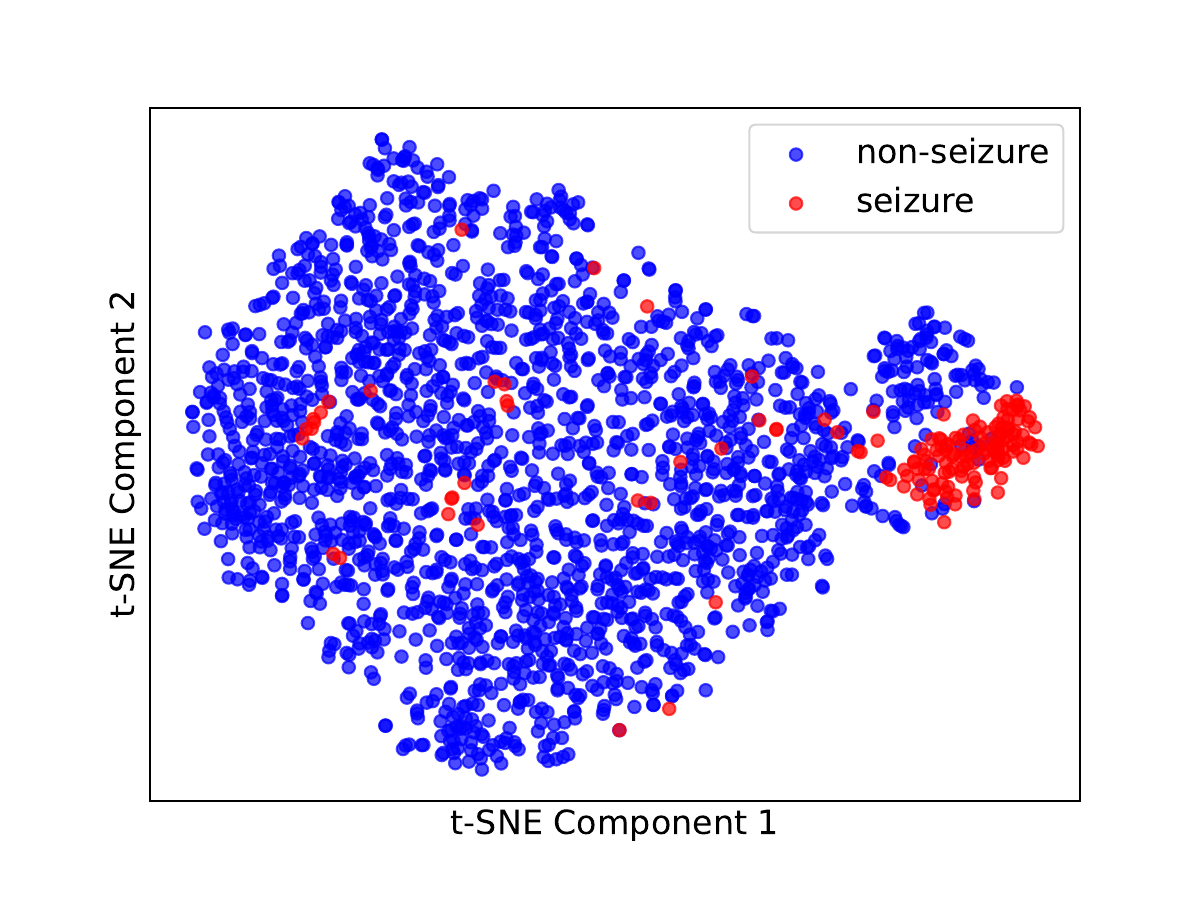}
        \caption{MR-EEGWaveNet \label{fig:feat_vis_mreegwavenet}}
    \end{subfigure}
    \caption{t-SNE visualization of features extracted from EEG recordings using EEGWaveNet and MR-EEGWaveNet (Dataset: Juntendo; Patient ID: Pt-$12$). \label{fig:feat_visual}}
\end{figure}
Furthermore, we used the t-distributed stochastic neighbor embedding (t-SNE) technique to visualize the distribution of features in two dimensions, as shown in Figure \ref{fig:feat_visual}. The features were extracted from the final layer before the prediction module in both the EEGWaveNet and the MR-EEGWaveNet models.

\subsection{Ablation Study with MR-EEGWaveNet Variants}
\label{sec:res_ablation_study}
Table~\ref{tab:ablation_compare_siena} presents the performance of two MR-EEGWaveNet variants and the baseline MR-EEGWaveNet-2 model on Siena dataset. Removing the feature extraction module (MR-EEGWaveNet-Abl1) significantly reduces performance, with a precision of $21.53\%$ and F1 score of $0.259$. MR-EEGWaveNet-Abl2, which excludes the $10$-second input stream, improves recall ($83.10\%$) but yields lower F1 ($0.288$) and AUC ($0.9364$) than the baseline. The baseline MR-EEGWaveNet-2 achieves the best performance across all metrics, including the highest precision ($26.74\%$), specificity ($93.86\%$) F1 score ($0.319$), and AUC ($0.9421$). The patient-wise performance details of the MR-EEGWaveNet-Abl1 and MR-EEGWaveNet-Abl2 models are reported in Appendix~\ref{appen:ptwise-results-siena}, in Tables~\ref{tab:MR-EEGWaveNet-Abl1-results} and~\ref{tab:MR-EEGWaveNet-Abl2-results}, respectively.

\begin{table}[t!]
\centering
\caption{Performance evaluation of MR-EEGWaveNet and ablated variants on the Siena dataset (best performances are highlighted in bold).}
\label{tab:ablation_compare_siena}
\scalebox{1.0}
{
\begin{tabular}{lccccc}
\hline
      \textbf{Models [Segments]}
    & \textbf{Pre.} 
    & \textbf{Rec.}	
    & \textbf{Spe.}	
    & \textbf{F1}
    & \textbf{AUC}
    \\\hline
    MR-EEGWaveNet-Abl1 {[}10 s{]}             & 21.53          & 81.23    & 91.75   & 0.259  & 0.9379 \\
    MR-EEGWaveNet-Abl2 {[}10 s{]}             & 23.17          & \textbf{83.10}    & 93.04   & 0.288  & 0.9364 \\
    \textbf{MR-EEGWaveNet-2 {[}10 s,  2 s{]}} & \textbf{26.74} & 82.39    & \textbf{93.86}   & \textbf{0.319}  & \textbf{0.9421} \\
    \hline
    
\end{tabular}
}
\end{table}

\begin{table}[t!]
\centering
\caption{Comparison of CBraMod (with LightGBM) and MR-EEGWaveNet performance on the Siena dataset.}
\label{tab:cbramod_compare}
\scalebox{1.0}
{
\begin{tabular}{lccccc}
\hline
      \textbf{Models [Segments]}
    & \textbf{Pre.} 
    & \textbf{Rec.}	
    & \textbf{Spe.}	
    & \textbf{F1}
    & \textbf{AUC}
    \\\hline
    CBraMod + LightGBM {[}10 s{]}    & 8.74      & 52.05    & 92.80   & 0.120  & 0.8756 \\
    MR-EEGWaveNet-2 {[}10 s,  2 s{]} & 26.74     & 82.39    & 93.86   & 0.319  & 0.9421 \\
    \hline
\end{tabular}
}
\end{table}

\subsection{Evaluation Against Pretrained Foundation Model}
\label{sec:sota_compare}

Table~\ref{tab:cbramod_compare} presents a performance comparison between the proposed MR-EEGWaveNet-2 model and the CBraMod model combined with LightGBM. The MR-EEGWaveNet-2 consistently outperforms the CBraMod-based approach across all evaluation metrics. Specifically, it achieves substantially higher precision ($26.74\%$ vs. $8.74\%$), recall ($82.39\%$ vs. $52.05\%$), F1 score ($0.319$ vs. $0.120$), and AUC ($0.9421$ vs. $0.8756$). The patient-wise performance of CBraMod combined with the LightGBM classifier is reported in Appendix~\ref{appen:ptwise-results-siena}, Table~\ref{tab:siena_ptwise_cbramod_results_without_pcp}.

\section{Discussion}
\label{sec:discussion}
\subsection{Effects of Window Length in Seizure Detection}
\label{sec:discuss-window-len}
The experimental results in Tables \ref{tab:comp_perf_siena_wo_preprocess} and \ref{tab:comp_perf_juntendo_wo_preprocess} indicate that the longer window tends to capture more characteristics of seizures than the shorter window. Moreover, it improves specificity in most cases, \ie, it efficiently separates noise, artifacts, and nonseizure EEG segments from seizure EEG. Based on our analysis and previous studies \cite{molla_hassan_graph_sz_det_sensor2020, hassan_epilepsy_CW_2019}, seizure EEG exhibits higher signal regularity compared to nonseizure EEG. With longer window, this regularity can be more effectively captured by the models. However, the longer window has less specific temporal information or features compared to a shorter window. Common artifacts, such as eye blink, eye movement, chewing, loose electrode, among others have some regularity, but their duration is significantly short, approximately $250$ ms to $3$ s \cite{islam_artifact_review_2016}. In contrast, some artifacts, such as loss of all electrode contact, muscle movement, or body movement, have long-duration effects on the EEG \cite{Kaya_summary_EEG_artifact_2021}. The experimental results indicate that the proposed architecture, MR-EEGWaveNet, effectively mitigates this issue. In addition, the EEG monitoring guidelines suggest a display of $10$ to $20$ seconds/page for routine EEG checkup \cite{EEG_guidlines_2016}.

\subsection{Effects of Post-Classification Processing}
\label{sec:post-class-process-effect}
An essential aspect of post-classification processing is selecting an appropriate threshold. Anomaly or outlier detection methods generally assign higher anomaly scores to artifacts, noise, or seizure EEG segments compared to background EEG activity \cite{hassan_ano_det_ICASSP2024, hassan_rimann_ATSIP_2024}. During seizures, synchronous neuronal activity across the whole brain or specific regions often causes the data distribution to deviate significantly from the mean anomaly score. Unless the recording is heavily contaminated with physiological or extra-physiological artifacts, the average anomaly score during seizures typically deviates significantly from the mean baseline. Although certain parts of a seizure may resemble normal EEG, they may be parts of the evolution of the seizure. Based on this observation, the mean anomaly score was chosen as the threshold, with experimental results supporting its effectiveness. However, more studies with larger datasets are necessary to determine the optimal threshold for real-world implementation.

\subsection{Performance Improvement with MR-EEGWaveNet}
\label{sec:discuss-perf-improv-mreegwavenet}
The experimental results on the Siena dataset (see Table \ref{tab:comp_perf_siena_wo_preprocess}) show that the MR-EEGWaveNet outperformed the EEGWaveNet model in most evaluation metrics, although it achieved a slightly lower recall. In particular, the MR-EEGWaveNet-2 and MR-EEGWaveNet-3 configurations achieved the highest performance, indicating a more robust detection capability. These results highlight the effectiveness of the MR-EEGWaveNet with multiresolution analysis compared to the existing EEGWaveNet model.
Similar to the Siena dataset, the experimental results on the Juntendo dataset (see Table \ref{tab:comp_perf_juntendo_wo_preprocess}) also demonstrate the superiority of the MR-EEGWaveNet model over the EEGWaveNet model. However, the recall of the MR-EEGWaveNet is slightly lower than that of the EEGWaveNet, but in these cases, the specificity of the EEGWaveNet is considerably lower, approximately $83\%$.

The patient-wise performance of the EEGWaveNet-3 and MR-EEGWaveNet-2 on the Siena dataset with post-classification processing is shown in Tables \ref{tab:siena_ptwise_eegwavenet_results_with_pcp} and \ref{tab:siena_ptwise_proposed_results_with_pcp}, respectively. Furthermore, Tables \ref{tab:junt_ptwise_eegwavenet_results_with_pcp} and \ref{tab:junt_ptwise_proposed_results_with_pcp} represent the results on the Juntendo dataset for the EEGWaveNet-3 and MR-EEGWaveNet-2 methods, respectively. Notably, the MR-EEGWaveNet demonstrates consistent performance across multiple datasets, further supporting its scalability and stability. However, more studies with larger datasets are necessary to confirm this finding. 

Additionally, Figure \ref{fig:feat_visual} illustrates that the features calculated by the MR-EEGWaveNet (Figure \ref{fig:feat_vis_mreegwavenet}) exhibit better class-wise separation than those produced by the EEGWaveNet (Figure \ref{fig:feat_vis_eegwavenet}), indicating an improved discriminative capability.

\subsection{Ablation Study on MR-EEGWaveNet Model Components}
\label{sec:discuss-ablation-study}
MR-EEGWaveNet-Abl1 excludes the feature extraction module, which significantly reduces the model's ability to learn rich temporal and spatial representations. This simplification leads to the lowest performance across all metrics (see Table \ref{tab:ablation_compare_siena})--indicating that the feature extractor plays a critical role in distinguishing seizure segments.

MR-EEGWaveNet-Abl2 removes the $10$-second input stream, retaining only the $2$-second sub-segment pipeline. This variant performs better than MR-EEGWaveNet-Abl1 in all metrics and achieves the highest recall ($83.10\%$), suggesting that the $2$-second sub-segments alone still provide useful information. However, the drop in other metrics, specifically AUC and specificity compared to the baseline, indicates that the longer context ($10$-second input) contributes to overall decision stability and discriminative power with fewer false positives.

The baseline model, MR-EEGWaveNet-2, which integrates both the feature extraction module and the multi-scale input pipeline (10 s and 2 s), achieves the best performance across all metrics. These improvements indicate that both components are complementary and essential to the model's effectiveness.

\subsection{Selection of MR-EEGWaveNet Model}
\label{sec:mr-eegwavenet-model-selection}
According to the performance of the proposed models in Table \ref{tab:comp_perf_siena_all} and \ref{tab:comp_perf_juntendo_all}, both MR-EEGWaveNet-2 {[}10 s,  2 s{]}, and MR-EEGWaveNet-3 {[}10 s, 5 s,  2 s{]} performs similar and better than MR-EEGWaveNet-1 {[}5 s, 2.5 s{]} in terms of precision, F1 score, and ROC-AUC. We conducted a Wilcoxon signed rank test using the Siena dataset to compare EEGWaveNet-3 with the MR-EEGWaveNet-2 and MR-EEGWaveNet-3 models, as all three models share the same window length. The p-value between EEGWaveNet-3 and MR-EEGWaveNet-2 is $0.0203$ $(<0.05)$, indicating a statistically significant performance difference, and MR-EEGWaveNet-2 outperforms EEGWaveNet-3 with a high degree of confidence. Furthermore, it indicates that the observed improvement is unlikely owing to random chance.

\subsection{Challenges in Comparing Seizure Detection Systems}
\label{sec:comparison_challenges}
Performance comparison between seizure detection systems is difficult owing to differences in datasets, evaluation metrics, and experimental setups. The experimental datasets vary in terms of patient counts, seizure types, and recording settings, making the results difficult to compare. In addition, evaluation strategies and performance metrics differ across studies. Experimental protocols also vary, for example, the patient-dependent vs. independent approach. The preprocessing steps vary; some studies use artifact and noise removal techniques to clean the data for further processing. The definition of seizure events can also vary between datasets. Differences in expert opinion may arise when classifying an EEG segment as a seizure, particularly at the onset and offset of the event. Furthermore, the lack of explainability of the implementation of the methods reduces the reproducibility of the published work. While certain systems support offline analysis, real-time applications introduce added constraints, particularly regarding latency. Moreover, some studies consider segment-based evaluation, while others are event-based.

\subsection{Comparison with Recent Studies}
\label{sec:comp_sota_sz_detection}
As shown in Table~\ref{tab:cbramod_compare}, the comparison with the CBraMod (with LightGBM) foundation model highlights the strengths of the proposed MR-EEGWaveNet-2 architecture in the context of seizure detection from long-term EEG recordings. Despite CBraMod's large-scale pre-training and use of powerful foundation model representations, its performance falls significantly short of MR-EEGWaveNet-2 when applied to the target dataset via feature extraction and LightGBM classification. Notably, MR-EEGWaveNet-2 achieved much higher precision ($26.74\%$ vs. $8.74\%$) and recall ($82.39\%$ vs. $52.05\%$), indicating its effectiveness in correctly identifying seizure events while minimizing false positives. Moreover, improvements in F1 score ($0.319$ vs. $0.120$) and AUC ($0.9421$ vs. $0.8756$) confirm the model's robustness against class imbalance and variability in EEG recordings. These findings indicate that end-to-end task-specific models like MR-EEGWaveNet outperform general pretrained models, especially in capturing domain-specific temporal and contextual features.

The Siena dataset utilized in this study is a relatively new dataset available publicly in this field. To the best of our knowledge, among the limited number of recent studies utilizing the Siena data set, the work by Sigsgaard et al. \cite{Sigsgaard2023_SZ_detection}, which adopts LOSO evaluation protocol, represents the most comparable baseline for evaluating our approach. Although the method proposed in \cite{Sigsgaard2023_SZ_detection} is segment-based, the performance evaluation reported in their study is event-based. 
Table \ref{tab:comp_perf_sota} presents a performance comparison between MR-EEGWaveNet-2 and the event-based approach proposed by Sigsgaard et al.
In terms of performance based on seizures, our proposed method outperforms the approach of Sigsgaard et al. \cite{Sigsgaard2023_SZ_detection} by approximately $13\%$ in the detection of seizures, with a seizure event detection rate of $\sim94\%$. However, in terms of specificity, it falls short by approximately $4\%$. Therefore, our proposed method effectively detects seizures in both segment-wise and event-based evaluation schemes, while maintaining a moderate level of specificity and a false positive rate. Furthermore, note that models designed explicitly for seizure event detection generally exhibit higher specificity compared to segment-based approaches. However, the performance reported largely depends on how seizure detection is defined and evaluated. In this study, a seizure event is considered detected if the model labels at least one EEG segment within the seizure period as a seizure, regardless of whether other segments of the seizure event are labeled. For example, if an annotated seizure occurs between 100 and 150 seconds and the EEG is divided into $10$-second segments, the event is considered detected if the model classifies any overlapping segment (e.g., 100--110 s or 110--120 s) as a seizure, even if not all segments within the 100--150 s period are detected.

\begin{table}[t!]
\centering
\caption{Performance comparison of the MR-EEGWaveNet with state-of-the-art works on the Siena dataset.}
\label{tab:comp_perf_sota}
\scalebox{1.0}
{
\begin{tabular}{lcccc}
\hline
      \textbf{Methods}
    & \textbf{Recall}
    & \textbf{Specificity}
    \\\hline
    Sigsgaard et al. \cite{Sigsgaard2023_SZ_detection}	     & 81.43	  & 99.38	 \\\hline
    MR-EEGWaveNet-2 & 94.24	  & 95.05	\\\hline
\end{tabular}
}
\end{table}

\subsection{Performance Metric Trade-offs in Seizure Detection}
\label{sec:metric-trade-off}
When designing a seizure detection model, selecting the optimal threshold between recall and specificity is crucial. Due to the highly imbalanced dataset, a single performance metric is insufficient to evaluate the model's performance. They could also be misleading if additional information about the dataset or patient is unavailable. Although recall is more important in seizure detection, an improvement in $2\%$ specificity despite a 2\% decrease in recall remains beneficial in terms of reducing the false alarm rate. This trade-off makes the classification system more reliable. In addition, evaluating the seizure detection rate in combination with other metrics is essential for a comprehensive assessment of performance. For segment-wise seizure detection, a recall in the range of $80$--$90\%$, combined with high specificity and a high rate of seizure event detection, indicates the suitability of a model for real-world applications. However, the model proposed in this study is specifically designed for offline seizure detection using long-term EEG recordings.

\subsection{Limitations and Future Directions}
\label{sec:limitation_future_dir}
A notable limitation of our model is its reduced performance for certain subjects. Further improvements are needed to enhance the model's ability to distinguish seizures from artifacts, such as loss of contact with the electrode, eye movements, or muscle activity. In the future, the model architecture should include context-based feature extraction to better capture temporal and spatial patterns. Furthermore, improving the model's overall reliability and performance will require the application of advanced artifact removal and post-classification processing techniques.

\section{Conclusion}
\label{sec:conclusion}
This study presents a novel end-to-end model to detect epileptic seizures from long EEG recordings. The length of the EEG segment plays a significant role in the model's performance, especially in distinguishing between seizure, noise, and artifact. The experimental results showed that the MR-EEGWaveNet with multiresolutional features outperforms existing end-to-end models in classifying seizure and nonseizure EEG signals. Additionally, we introduced an anomaly score-based post-classification processing technique to improve performance and reduce false positive rates. We also evaluated the performance of MR-EEGWaveNet under different parameter settings and with ablated versions of its architecture. Despite lower performance in certain subjects, the improvement achieved with MR-EEGWaveNet indicates its potential for continued development. However, enhancing MR-EEGWaveNet is necessary to better distinguish between artifacts and seizure events. Although the MR-EEGWaveNet was tested on two datasets and demonstrated consistent performance, further investigation with larger datasets is required to apply it in real-world applications. 

\section*{Code Availability}
The code used in this study is available at \url{https://github.com/ttlabtuat/MR-EEGWaveNet}.

\clearpage
\appendix

\section{Ablated Variants of MR-EEGWaveNet}
\label{appen:ablation_model}
\setcounter{table}{0}
\renewcommand{\thetable}{A\arabic{table}}

\setcounter{figure}{0}
\renewcommand{\thefigure}{A\arabic{figure}}

\subsection{Architecture of MR-EEGWaveNet Ablation Model 1 (MR-EEGWaveNet-Abl1)}
\label{appen:abl_model_1}
\begin{figure}[th!]
    \centering
    \centerline{\includegraphics[width=0.9\linewidth]{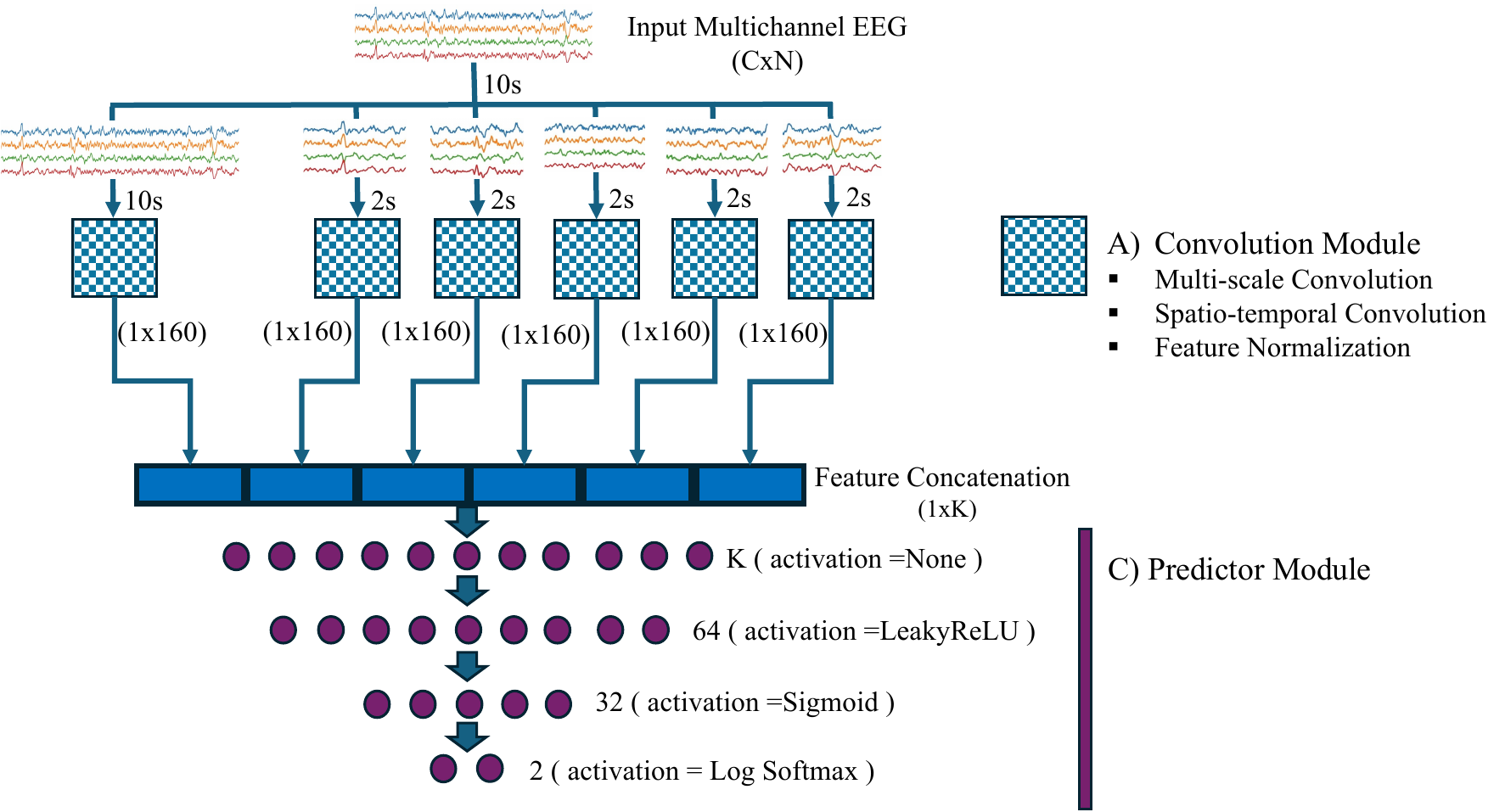}}
    \caption{MR-EEGWaveNet-Abl1 architecture}
    \label{fig:mr-eegwavenet-abl1-arch-image}
\end{figure}

\subsection{Architecture of MR-EEGWaveNet Ablation Model 2 (MR-EEGWaveNet-Abl2)}
\label{appen:abl_model_2}
\begin{figure}[th!]
    \centering
    \centerline{\includegraphics[width=0.9\linewidth]{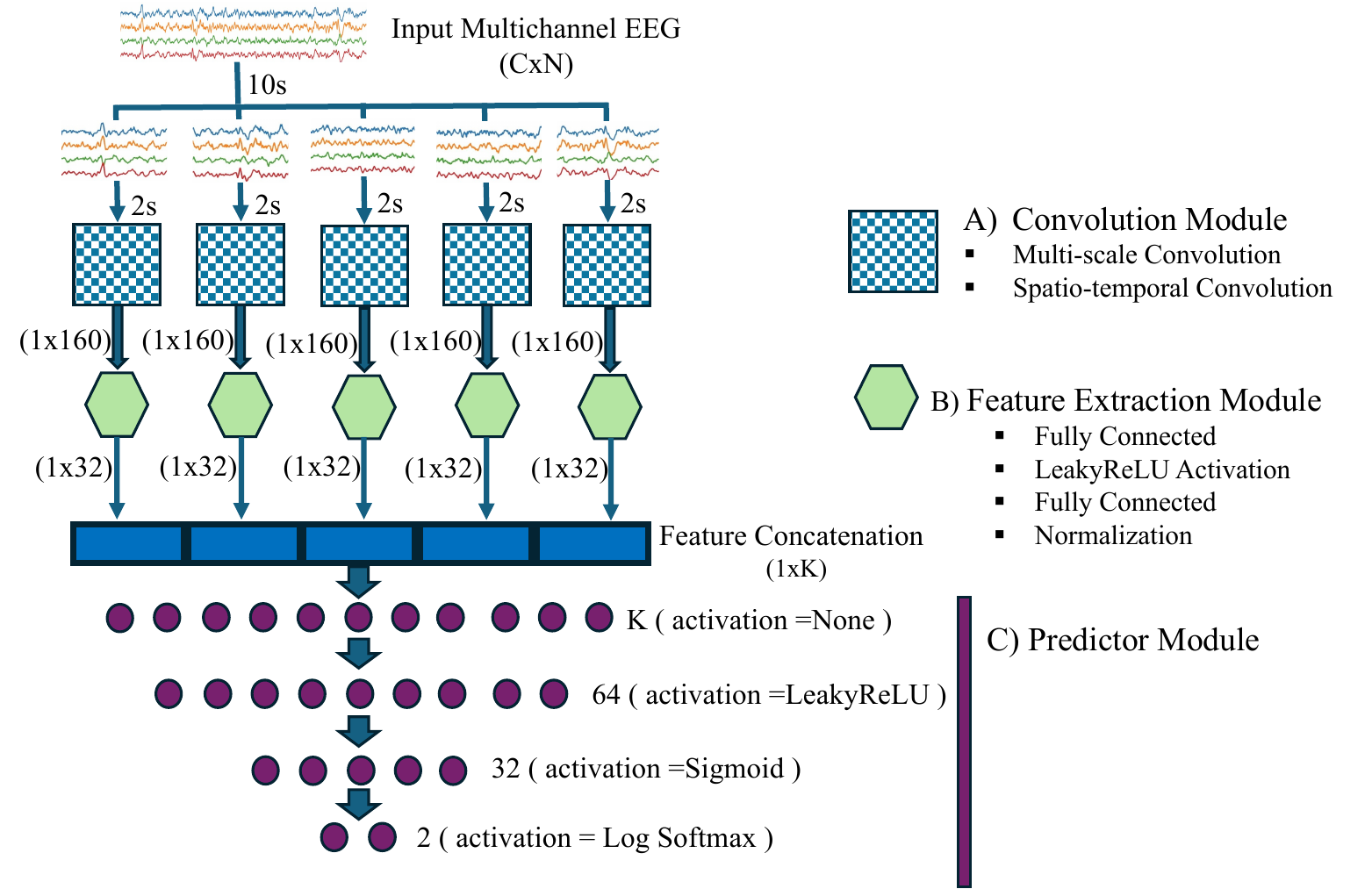}}
    \caption{MR-EEGWaveNet-Abl2 architecture}
    \label{fig:mr-eegwavenet-abl2-arch-image}
\end{figure}

\clearpage
\section{Siena Dataset: Patient-Wise Results}
\label{appen:ptwise-results-siena}

\setcounter{table}{0}
\renewcommand{\thetable}{B\arabic{table}}

\subsection{EEGWaveNet Model}
\label{appen:pt-results-eegwavenet-siena}

\begin{table}[!ht]
\centering
\caption{EEGWaveNet-3, Patient-wise performance (Window: $10$ s), with post-classification processing on Siena dataset.}
\label{tab:siena_ptwise_eegwavenet_results_with_pcp}
\scalebox{0.95}
{
\begin{tabular}{c|r|r|r|r|r|c}
\hline
\textbf{Pt.ID} & \textbf{Pre.} & \textbf{Rec.} & \textbf{Spe.} & \textbf{F1} & \textbf{AUC} & \textbf{Det. Ratio} \\ \hline
        PN00 & 45.25 & 100.0  & 96.95 & 0.622 & 0.996 & 1.000 \\ \hline
        PN01 & 3.92  & 100.0  & 87.62 & 0.074 & 0.970 & 1.000 \\ \hline
        PN03 & 1.71  & 86.07  & 87.47 & 0.033 & 0.881 & 1.000 \\ \hline
        PN05 & 1.42  & 58.52  & 81.66 & 0.028 & 0.903 & 0.889 \\ \hline
        PN06 & 18.55 & 80.77  & 96.52 & 0.285 & 0.967 & 0.867 \\ \hline
        PN07 & 0.95  & 80.00  & 84.82 & 0.019 & 0.907 & 1.000 \\ \hline
        PN09 & 44.73 & 83.33  & 95.57 & 0.406 & 0.975 & 0.889 \\ \hline
        PN10 & 17.04 & 35.32  & 98.40 & 0.201 & 0.683 & 0.407 \\ \hline
        PN11 & 5.24  & 94.44  & 87.74 & 0.098 & 0.914 & 1.000 \\ \hline
        PN12 & 3.74  & 91.22  & 70.81 & 0.072 & 0.901 & 1.000 \\ \hline
        PN13 & 10.74 & 92.06  & 93.52 & 0.191 & 0.958 & 1.000 \\ \hline
        PN14 & 2.78  & 69.05  & 95.19 & 0.053 & 0.911 & 0.750 \\ \hline
        PN16 & 6.71  & 89.39  & 83.08 & 0.124 & 0.917 & 1.000 \\ \hline
        PN17 & 16.43 & 92.68  & 95.99 & 0.276 & 0.942 & 1.000 \\ \hline
\textbf{Mean} & \textbf{12.80} & \textbf{82.35} & \textbf{89.67} & \textbf{0.177} & \textbf{0.918} & \textbf{0.914} \\ \hline
\end{tabular}
}
\end{table}

\blfootnote{Pt.ID: Patient ID; Pre.: Precision; Rec.: Recall; Spe.: Specificity; F1: F1 score; The performance for each metric is reported as the average of three runs per patient.}

\subsection{MR-EEGWaveNet: Multiresolutional EEGWaveNet}
\label{appen:pt-results-proposed-siena}

\begin{table}[!ht]
\centering
\caption{MR-EEGWaveNet-2, Patient-wise performance (Window: $10$ s), with post-classification processing on Siena dataset.}
\label{tab:siena_ptwise_proposed_results_with_pcp}
\scalebox{0.95}
{
\begin{tabular}{c|r|r|r|r|r|c}
\hline
\textbf{Pt.ID} & \textbf{Pre.} & \textbf{Rec.} & \textbf{Spe.} & \textbf{F1} & \textbf{AUC} & \textbf{Det. Ratio} \\ \hline
        PN00 & 86.96 & 91.03 & 99.68 & 0.888 & 0.998 & 1.000 \\ \hline
        PN01 & 30.89 & 100.0 & 98.11 & 0.405 & 0.999 & 1.000 \\ \hline
        PN03 & 2.39  & 90.91 & 90.30 & 0.047 & 0.952 & 1.000 \\ \hline
        PN05 & 2.15  & 55.56 & 89.35 & 0.041 & 0.966 & 1.000 \\ \hline
        PN06 & 47.09 & 76.25 & 99.52 & 0.581 & 0.991 & 1.000 \\ \hline
        PN07 & 2.05  & 87.78 & 92.24 & 0.040 & 0.963 & 1.000 \\ \hline
        PN09 & 73.78 & 88.41 & 99.81 & 0.801 & 0.995 & 1.000 \\ \hline
        PN10 & 13.02 & 40.07 & 97.50 & 0.162 & 0.778 & 0.444 \\ \hline
        PN11 & 6.85  & 83.33 & 90.28 & 0.122 & 0.827 & 1.000 \\ \hline
        PN12 & 9.76  & 89.59 & 89.04 & 0.175 & 0.945 & 1.000 \\ \hline
        PN13 & 13.88 & 94.32 & 95.32 & 0.241 & 0.981 & 1.000 \\ \hline
        PN14 & 4.70  & 62.27 & 97.24 & 0.086 & 0.918 & 0.750 \\ \hline
        PN16 & 11.03 & 68.18 & 92.33 & 0.187 & 0.912 & 1.000 \\ \hline
        PN17 & 97.22 & 87.55 & 99.98 & 0.921 & 0.965 & 1.000 \\ \hline
\textbf{Mean} & \textbf{28.70} & \textbf{79.66} & \textbf{95.05} & \textbf{0.336} & \textbf{0.942} & \textbf{0.942} \\ \hline
\end{tabular}
}
\end{table}

\clearpage
\subsection{MR-EEGWaveNet Ablation Model 1 (MR-EEGWaveNet-Abl1)}
\label{appen:MR-EEGWaveNet-Abl1_results}
\begin{table}[!ht]
\centering
\caption{MR-EEGWaveNet-Abl1, Patient-wise performance (Window: $10$ s), without post-classification processing on Siena dataset.}
\label{tab:MR-EEGWaveNet-Abl1-results}

\scalebox{0.95}{
\begin{tabular}{c|r|r|r|r|r|c}
\hline

\textbf{Pt.ID} & \textbf{Pre.} & \textbf{Rec.} & \textbf{Spe.} & \textbf{F1} & \textbf{AUC} & \textbf{Det. Ratio} \\ \hline
PN00 & 77.27 & 98.72 & 98.94 & 0.850 & 0.999 & 1.000 \\ \hline
PN01 & 7.80  & 100.0 & 96.43 & 0.144 & 1.000 & 1.000 \\ \hline
PN03 & 2.53  & 92.42 & 90.06 & 0.049 & 0.970 & 1.000 \\ \hline
PN05 & 4.35  & 87.96 & 88.96 & 0.082 & 0.951 & 1.000 \\ \hline
PN06 & 28.67 & 80.29 & 97.71 & 0.358 & 0.986 & 1.000 \\ \hline
PN07 & 1.58  & 86.67 & 90.57 & 0.031 & 0.952 & 1.000 \\ \hline
PN09 & 86.06 & 83.65 & 99.92 & 0.847 & 0.998 & 1.000 \\ \hline
PN10 & 10.06 & 46.20 & 98.15 & 0.161 & 0.891 & 0.482 \\ \hline
PN11 & 3.41  & 63.89 & 85.62 & 0.063 & 0.777 & 1.000 \\ \hline
PN12 & 3.09  & 90.56 & 60.13 & 0.060 & 0.904 & 1.000 \\ \hline
PN13 & 10.66 & 97.22 & 92.92 & 0.191 & 0.977 & 1.000 \\ \hline
PN14 & 6.60  & 63.55 & 98.19 & 0.119 & 0.915 & 0.667 \\ \hline
PN16 & 15.70 & 61.23 & 89.32 & 0.216 & 0.882 & 1.000 \\ \hline
PN17 & 43.62 & 84.80 & 97.60 & 0.465 & 0.930 & 1.000 \\ \hline
\textbf{Mean} & \textbf{21.53} & \textbf{81.23} & \textbf{91.75} & \textbf{0.259} & \textbf{0.937} & \textbf{0.939} \\ \hline
\end{tabular}
}
\end{table}

\subsection{MR-EEGWaveNet Ablation Model 2 (MR-EEGWaveNet-Abl2)}
\label{appen:MR-EEGWaveNet-Abl2_results}
\begin{table}[!ht]
\centering
\caption{MR-EEGWaveNet-Abl2, Patient-wise performance (Window: $10$ s), without post-classification processing on Siena dataset.}
\label{tab:MR-EEGWaveNet-Abl2-results}
\scalebox{0.95}{
\begin{tabular}{c|r|r|r|r|r|c}
\hline
\textbf{Pt.ID} & \textbf{Pre.} & \textbf{Rec.} & \textbf{Spe.} & \textbf{F1} & \textbf{AUC} & \textbf{Det. Ratio} \\ \hline
PN00 & 72.49 & 100.0 & 99.13 & 0.840 & 0.999 & 1.000 \\ \hline
PN01 & 16.69 & 100.0 & 96.26 & 0.269 & 1.000 & 1.000 \\ \hline
PN03 & 2.23  & 91.17 & 88.60 & 0.044 & 0.945 & 1.000 \\ \hline
PN05 & 3.25  & 96.30 & 87.40 & 0.063 & 0.946 & 1.000 \\ \hline
PN06 & 34.19 & 58.75 & 99.37 & 0.409 & 0.971 & 0.933 \\ \hline
PN07 & 1.13  & 93.33 & 85.85 & 0.022 & 0.927 & 1.000 \\ \hline
PN09 & 47.95 & 81.11 & 97.89 & 0.520 & 0.988 & 1.000 \\ \hline
PN10 & 13.08 & 45.08 & 98.64 & 0.193 & 0.799 & 0.593 \\ \hline
PN11 & 4.27  & 100.0 & 83.93 & 0.082 & 0.934 & 1.000 \\ \hline
PN12 & 6.19  & 87.96 & 80.51 & 0.115 & 0.921 & 1.000 \\ \hline
PN13 & 23.89 & 85.04 & 96.61 & 0.346 & 0.975 & 1.000 \\ \hline
PN14 & 5.55  & 61.17 & 97.88 & 0.101 & 0.899 & 0.667 \\ \hline
PN16 & 9.02  & 71.21 & 90.69 & 0.160 & 0.873 & 1.000 \\ \hline
PN17 & 84.47 & 92.28 & 99.87 & 0.881 & 0.933 & 1.000 \\ \hline
\textbf{Mean} & \textbf{23.17} & \textbf{83.10} & \textbf{93.04} & \textbf{0.288} & \textbf{0.936} & \textbf{0.942} \\ \hline
\end{tabular}
}
\end{table}

\blfootnote{Pt.ID: Patient ID; Pre.: Precision; Rec.: Recall; Spe.: Specificity; F1: F1 score; The performance for each metric is reported as the average of three runs per patient.}

\clearpage
\subsection{CBraMod: Criss-Cross Brain Foundation Model}
\label{appen:pt-results-cbramod-siena}

\begin{table}[!ht]
\centering
\caption{CBraMod+LightGBM, Patient-wise performance (Window: $10$ s), without post-classification processing on Siena dataset.}
\label{tab:siena_ptwise_cbramod_results_without_pcp}
\scalebox{0.95}{
\begin{tabular}{c|r|r|r|r|r|c}
\hline
\textbf{Pt.ID} & \textbf{Pre.} & \textbf{Rec.} & \textbf{Spe.} & \textbf{F1} & \textbf{AUC} & \textbf{Det. Ratio} \\ \hline
        PN00 & 33.29 & 35.02 & 98.45 & 0.339 & 0.897 & 0.867 \\ \hline
        PN01 & 6.16  & 48.48 & 98.41 & 0.109 & 0.949 & 1.000 \\ \hline
        PN03 & 1.90  & 96.97 & 87.27 & 0.037 & 0.980 & 1.000 \\ \hline
        PN05 & 0.45  & 22.22 & 79.74 & 0.009 & 0.631 & 0.556 \\ \hline
        PN06 & 3.22  & 12.70 & 98.06 & 0.051 & 0.808 & 0.400 \\ \hline
        PN07 & 0.97  & 100.0 & 80.51 & 0.019 & 0.923 & 1.000 \\ \hline
        PN09 & 5.50  & 20.16 & 97.92 & 0.086 & 0.850 & 0.556 \\ \hline
        PN10 & 24.48 & 30.21 & 99.64 & 0.270 & 0.888 & 0.296 \\ \hline
        PN11 & 5.18  & 93.33 & 90.06 & 0.098 & 0.935 & 1.000 \\ \hline
        PN12 & 6.99  & 57.55 & 91.72 & 0.125 & 0.872 & 1.000 \\ \hline
        PN13 & 2.78  & 17.10 & 95.69 & 0.048 & 0.828 & 0.667 \\ \hline
        PN14 & 3.79  & 53.85 & 97.57 & 0.071 & 0.835 & 0.500 \\ \hline
        PN16 & 7.26  & 89.39 & 85.47 & 0.134 & 0.951 & 1.000 \\ \hline
        PN17 & 20.32 & 51.71 & 98.63 & 0.289 & 0.913 & 1.000 \\ \hline
\textbf{Mean} & \textbf{8.74} & \textbf{52.05} & \textbf{92.80} & \textbf{0.120} & \textbf{0.875} & \textbf{0.774} \\ \hline
\end{tabular}
}
\end{table}

\blfootnote{Pt.ID: Patient ID; Pre.: Precision; Rec.: Recall; Spe.: Specificity; F1: F1 score; The performance for each metric is reported as the average of three runs per patient.}

\clearpage
\section{Juntendo Dataset: Patient-wise Results}
\label{appen:ptwise-results-juntendo}

\setcounter{table}{0}
\renewcommand{\thetable}{C\arabic{table}}

\subsection{EEGWaveNet Model}
\label{appen:pt-results-eegwavenet-junt}

\begin{table}[!ht]
\centering
\caption{EEGWaveNet-3, Patient-wise performance (Window: $10$ s), with post-classification processing on Juntendo dataset.}
\label{tab:junt_ptwise_eegwavenet_results_with_pcp}
\scalebox{1.0}{
\begin{tabular}{c|r|r|r|r|r|c}
\hline
\textbf{Pt.ID} & \textbf{Pre.} & \textbf{Rec.} & \textbf{Spe.} & \textbf{F1} & \textbf{AUC} & \textbf{Det. Ratio} \\ \hline
        Pt-01 & 50.27  & 93.65  & 98.10  & 0.633  & 0.995  & 0.922 \\ \hline
        Pt-02 & 4.35   & 100.00 & 66.32  & 0.083  & 0.909  & 1.000 \\ \hline
        Pt-03 & 20.62  & 99.05  & 94.30  & 0.335  & 0.989  & 1.000 \\ \hline
        Pt-04 & 5.22   & 100.00 & 91.55  & 0.098  & 0.997  & 1.000 \\ \hline
        Pt-05 & 6.81   & 83.47  & 81.45  & 0.126  & 0.798  & 1.000 \\ \hline
        Pt-06 & 37.42  & 98.29  & 96.13  & 0.529  & 0.995  & 1.000 \\ \hline
        Pt-07 & 31.21  & 97.92  & 96.25  & 0.454  & 0.992  & 1.000 \\ \hline
        Pt-08 & 15.02  & 95.38  & 90.90  & 0.258  & 0.956  & 1.000 \\ \hline
        Pt-09 & 74.07  & 83.55  & 94.17  & 0.779  & 0.887  & 1.000 \\ \hline
        Pt-10 & 29.51  & 74.20  & 98.97  & 0.417  & 0.911  & 1.000 \\ \hline
        Pt-11 & 25.51  & 80.81  & 94.27  & 0.384  & 0.960  & 1.000 \\ \hline
        Pt-12 & 46.65  & 66.11  & 93.81  & 0.529  & 0.811  & 1.000 \\ \hline
        Pt-13 & 5.99   & 85.71  & 72.19  & 0.112  & 0.906  & 1.000 \\ \hline
        Pt-14 & 23.58  & 86.20  & 96.39  & 0.352  & 0.989  & 1.000 \\ \hline
        Pt-15 & 4.62   & 97.44  & 81.88  & 0.088  & 0.826  & 1.000 \\ \hline
        Pt-16 & 49.92  & 96.30  & 98.45  & 0.644  & 0.998  & 1.000 \\ \hline
        Pt-17 & 2.88   & 66.67  & 90.44  & 0.053  & 0.895  & 0.667 \\ \hline
        Pt-18 & 2.91   & 76.67  & 87.89  & 0.056  & 0.870  & 1.000 \\ \hline
        Pt-19 & 22.44  & 96.40  & 95.34  & 0.356  & 0.974  & 0.957 \\ \hline
        Pt-20 & 9.30   & 68.23  & 91.64  & 0.163  & 0.814  & 1.000 \\ \hline
        Pt-21 & 26.45  & 100.00 & 99.21  & 0.417  & 0.995  & 1.000 \\ \hline
\textbf{Mean} & \textbf{23.56} & \textbf{87.91}  & \textbf{90.94} & \textbf{0.327} & \textbf{0.927} & \textbf{0.978} \\ \hline
\end{tabular}
}
\end{table}

\blfootnote{Pt.ID: Patient ID; Pre.: Precision; Rec.: Recall; Spe.: Specificity; F1: F1 score; The performance for each metric is reported as the average of three runs per patient.}

\clearpage
\subsection{MR-EEGWaveNet: Multiresolutional EEGWaveNet}
\label{appen:pt-results-propsoed-junt}

\begin{table}[!ht]
\centering
\caption{MR-EEGWaveNet-2, Patient-wise performance (Window: $10$ s), with post-classification processing on Juntendo dataset.}
\label{tab:junt_ptwise_proposed_results_with_pcp}
\scalebox{1.0}{
\begin{tabular}{c|r|r|r|r|r|c}
\hline
\textbf{Pt.ID} & \textbf{Pre.} & \textbf{Rec.} & \textbf{Spe.} & \textbf{F1} & \textbf{AUC} & \textbf{Det. Ratio} \\ \hline
        Pt-01 & 71.99 & 83.57 & 99.46 & 0.765 & 0.993 & 0.804 \\ \hline
        Pt-02 & 4.31  & 92.28 & 68.67 & 0.082 & 0.886 & 1.000 \\ \hline
        Pt-03 & 44.98 & 94.20 & 98.42 & 0.601 & 0.991 & 1.000 \\ \hline
        Pt-04 & 29.52 & 100.00 & 99.26 & 0.450 & 0.999 & 1.000 \\ \hline
        Pt-05 & 43.78 & 55.81 & 98.75 & 0.473 & 0.919 & 0.556 \\ \hline
        Pt-06 & 51.22 & 90.77 & 98.24 & 0.647 & 0.983 & 1.000 \\ \hline
        Pt-07 & 74.21 & 97.92 & 99.63 & 0.844 & 0.998 & 1.000 \\ \hline
        Pt-08 & 18.91 & 88.89 & 93.89 & 0.312 & 0.958 & 1.000 \\ \hline
        Pt-09 & 95.63 & 79.45 & 99.28 & 0.866 & 0.970 & 0.889 \\ \hline
        Pt-10 & 67.17 & 44.01 & 99.90 & 0.532 & 0.986 & 0.867 \\ \hline
        Pt-11 & 49.08 & 78.12 & 97.99 & 0.593 & 0.981 & 1.000 \\ \hline
        Pt-12 & 50.13 & 51.43 & 95.83 & 0.487 & 0.877 & 1.000 \\ \hline
        Pt-13 & 4.65  & 76.83 & 71.00 & 0.088 & 0.835 & 1.000 \\ \hline
        Pt-14 & 34.47 & 70.35 & 98.32 & 0.423 & 0.986 & 1.000 \\ \hline
        Pt-15 & 5.40  & 94.44 & 85.15 & 0.102 & 0.816 & 1.000 \\ \hline
        Pt-16 & 96.67 & 100.00 & 99.95 & 0.982 & 0.996 & 1.000 \\ \hline
        Pt-17 & 2.44  & 50.00 & 91.45 & 0.046 & 0.780 & 0.500 \\ \hline
        Pt-18 & 2.84  & 46.67 & 94.22 & 0.054 & 0.902 & 0.500 \\ \hline
        Pt-19 & 76.00 & 74.81 & 99.63 & 0.724 & 0.989 & 0.731 \\ \hline
        Pt-20 & 15.38 & 59.34 & 95.83 & 0.240 & 0.920 & 1.000 \\ \hline
        Pt-21 & 88.89 & 100.00 & 99.95 & 0.933 & 0.998 & 1.000 \\ \hline
\textbf{Mean} & \textbf{44.18} & \textbf{77.57} & \textbf{94.51} & \textbf{0.488} & \textbf{0.941} & \textbf{0.897} \\ \hline
\end{tabular}
}
\end{table}

\blfootnote{Pt.ID: Patient ID; Pre.: Precision; Rec.: Recall; Spe.: Specificity; F1: F1 score; The performance for each metric is reported as the average of three runs per patient.}

\clearpage
\section{LightGBM Classifier}
\label{appen:lightGBM}

\setcounter{table}{0}
\renewcommand{\thetable}{D\arabic{table}}

\begin{table}[htbp!]
\centering
\caption{LightGBM parameters for training }
\label{tab:lightGBM_default_param}
\begin{tabular}{ll}
\hline
\textbf{Parameter} & \textbf{Value} \\
\hline
    boosting\_type       & gbdt \\
    class\_weight        & balanced \\
    colsample\_bytree    & 1.0 \\
    importance\_type     & split \\
    learning\_rate       & 0.1 \\
    max\_depth           & -1 \\
    min\_child\_samples  & 20 \\
    min\_child\_weight   & 0.001 \\
    min\_split\_gain     & 0.0 \\
    n\_estimators        & 100 \\
    n\_jobs              & None \\
    num\_leaves          & 31 \\
    objective            & None \\
    random\_state        & 42 \\
    reg\_alpha           & 0.0 \\
    reg\_lambda          & 0.0 \\
    subsample            & 1.0 \\
    subsample\_for\_bin  & 200000 \\
    subsample\_freq      & 0 \\
\hline
\end{tabular}
\end{table}

\clearpage
\bibliographystyle{ieeetr}
\bibliography{reference}
\end{document}